\documentclass[journal]{IEEEtran}

\usepackage{xcolor,soul,framed} 
\colorlet{shadecolor}{yellow}
\usepackage[pdftex]{graphicx}
\graphicspath{{../pdf/}{../jpeg/}}
\DeclareGraphicsExtensions{.pdf,.jpeg,.png}
\usepackage[cmex10]{amsmath}
\usepackage{array}
\usepackage{mdwmath}
\usepackage{mdwtab}
\usepackage{eqparbox}
\usepackage{url}
\usepackage{floatrow}
\usepackage[colorlinks]{hyperref}
\usepackage[colorinlistoftodos]{todonotes}

\usepackage[linesnumbered,ruled,vlined]{algorithm2e}
\usepackage{graphicx,caption}
\usepackage{eso-pic}
\usepackage{lipsum}

\usepackage{array}
\usepackage{mdwmath}
\usepackage{mdwtab}
\usepackage{eqparbox}
\usepackage{url}
\usepackage{floatrow}
\usepackage{setspace} 
\usepackage{adjustbox}
\usepackage{boldline}
\usepackage{graphicx}
\usepackage{subcaption}	
\usepackage{caption}
\usepackage{mathtools}
\usepackage{array, rotating}
\usepackage{multirow, makecell}
\usepackage{amssymb}
\usepackage[flushleft]{threeparttable}
\usepackage[linesnumbered,ruled,vlined]{algorithm2e}

\SetKwInput{KwInput}{Input}                
\SetKwInput{KwOutput}{Output}  
\newcommand\norm[1]{\lVert#1\rVert}
\usepackage[switch]{lineno}

\begin{document}
\AddToShipoutPictureBG*{%
  \AtPageUpperLeft{%
    \hspace{\paperwidth}%
    \raisebox{-\baselineskip}{%
    \footnotesize
        \makebox[0pt][r]{This work has been submitted to the IEEE for possible publication. Copyright may be transferred without notice, after which this version may no longer be accessible.~~~~~~~}\\
}}}%

\bstctlcite{IEEEexample:BSTcontrol}
    \title{A Novel Skeleton-Based Human Activity Discovery Using Particle Swarm Optimization with Gaussian Mutation}
  \author{Parham Hadikhani*, Daphne Teck Ching Lai and Wee-Hong Ong
 
  \thanks{P. Hadikhani, D.T.C. Lai, and~ W.H. Ong are with the School of Digital Science, Universiti Brunei Darussalam, Brunei. E-mail: \{20h8561, daphne.lai, weehong.ong\}@ubd.edu.bn.}
 }  


\maketitle

\begin{abstract}
Human activity discovery aims to cluster the activities performed by humans without any prior information on what defines each activity. Most methods presented in human activity recognition are supervised, where there are labeled inputs to train the system. In reality, it is difficult to label activities data because of its huge volume and the variety of human activities. This paper proposes an unsupervised framework to perform human activity discovery in 3D skeleton sequences. First, an approach for data pre-processing is presented. In this stage, important frames are selected based on kinetic energy. Next, the displacement of joints, statistical displacements, angles, and orientation features are extracted to represent the activities information. Since not all extracted features have useful information, the dimension of features is reduced using PCA. Most methods proposed for human activity discovery are not fully unsupervised. They use pre-segmented videos before categorizing activities. To deal with this, we have used a sliding time window to segment the time series of activities with some overlapping. Then, activities are discovered by our proposed Hybrid Particle swarm optimization (PSO) with Gaussian Mutation and K-means (HPGMK) algorithm to provide diverse solutions. PSO is used
due to its straightforward idea and powerful global search capability which can identify the ideal solution in a few iterations. Finally, k-means is applied to the outcome centroids from each iteration of the PSO to overcome the slow convergence rate of PSO. The experiment results on five datasets show that the proposed framework has superior performance in discovering activities compared to the other state-of-the-art methods and has increased accuracy of at least 4\,\% on average.
\end{abstract}


\IEEEpeerreviewmaketitle

 \section{Introduction}\label{sec1}
\begin{figure}
 \begin{center}
  \includegraphics[height=2.5 in ,width=\textwidth]{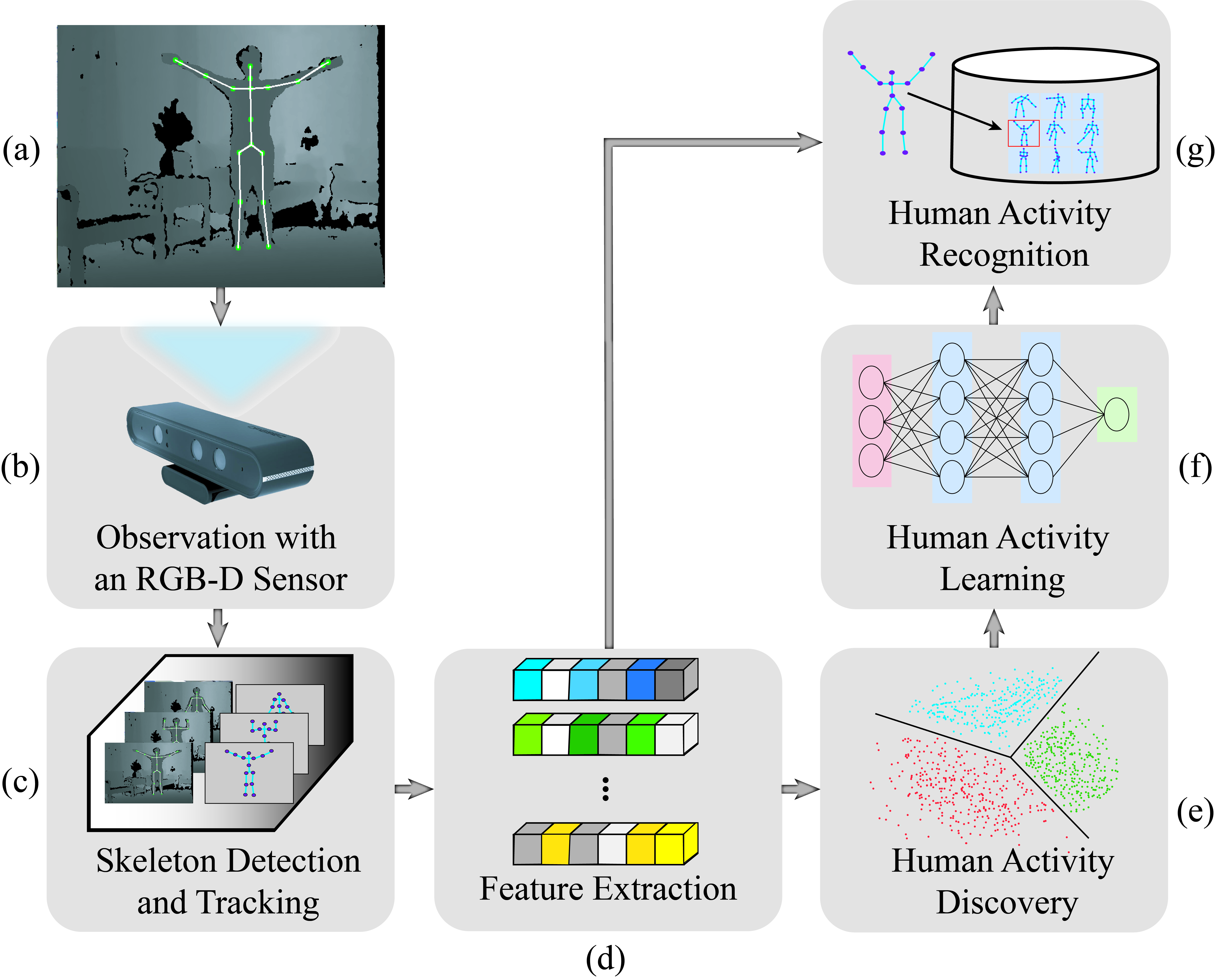}
  \caption{Overview of HAR system: (a) performed activities are (b) captured by a Kinect sensor. (c) After that, pose of humans are estimated by extracting joints. (d) To make raw data more usable, their salient and defining features are identified. (e) Based on the similarities and differences, activities are discovered. (f) Afterwards, the system begins to learn from the discovered activities and (g) finally human activities are recognized. }\label{conceptual overview}
  \end{center}
\end{figure}

\IEEEPARstart{H}{uman} Activity Recognition (HAR) has attracted much attention due to its applications in fields such as human-computer interaction, intelligent transportation systems, and monitoring applications ~\cite{chandrashekhar2006human}. Activity recognition aims to identify actions and activities that humans perform in different environments automatically. The input to a vision-based HAR system is a sequence of frames of a person performing different movements. The output is a set of labels representing the actions taken or activities in those movements. Many existing works use visual data as input. But such data have considerable complexity detrimental to the performance of HAR systems. These complexities include cluttered background, changes in brightness and points of view. 3D skeleton data partially overcomes these complexities and protects people's privacy when RGB data is not captured. Each frame represented by 3D coordinates of the main body joints is appropriate for representing human actions ~\cite{paoletti2021subspace} and can be easily obtained in real-time with low-cost depth sensors ~\cite{han2013enhanced}.\\
As shown in Fig.~\ref{conceptual overview}, there are at least seven steps in vision-based HAR systems. Vision sensors capture activities performed by a person. The skeletal information comprising joints coordinate are extracted from captured videos, containing image sequences called frames. Meaningful features are then extracted for more accurate activity discovery. The system without using manual annotations and having any guidance for activities discovers them by clustering the most similar activities from a set of different activities. In other words, the system tries to differentiate observed activities based on the likeness of extracted features. The discovered activity clusters are used in the learning process to model each cluster of activity and recognize future activities. \\
Significant progress has been made in the supervised learning of activity models~\cite{su2020predict}, illustrated in blocks (f) and (g) of  Fig.~\ref{conceptual overview}.~The learning and recognition steps rely on human-labeled training data to categorize activities if activity discovery (block e) was not performed. Human activity discovery is a part of the HAR process where activities are categorized based on their similarities without any knowledge of activity labels or any information that characterizes an activity, making this step particularly challenging. In other words,  activity discovery is like a child's learning. There is no prior information to define a specific sequence of movements to mean a particular activity such as crawling or waving and so forth to the child learner. Using the ability to differentiate, they learn from unlabeled data and form a model that can post-label new data based on that training. In human activity discovery, there is no known information or knowledge about a particular movement, including its start to end, for example when someone is picking up something. This means the input is a series of movements without knowing the start and endpoints to indicate each activity. Some existing work has segmented the input data by activity~\cite{paoletti2021subspace}. Thus, the start and endpoints of the activities are already known, although the method of grouping activities may be unsupervised. \\
In this paper, we focus on the less developed activity discovery comprising the block (d) to (e) of Fig.~\ref{conceptual overview} by developing an effective methodology to extract good features and cluster activities without any label. Keyframe selection~\cite{shan20143d} and PCA are used to remove redundant frames and features to reduce time complexity and increase accuracy.A  pre-processing and feature extraction methodology are proposed to prepare information and extract features from the most informative joints and bones, including joint displacement, joint orientation, and statistical time domain. As our first study, a hybrid PSO with Gaussian Mutation and k-means (HPGMK) clustering that requires a known prior of the cluster number is proposed to find activities. Sometimes particles converge to a specific point between the best global and personal positions and get trapped in local optima. This difficulty arises when the swarm's variety reduces and the swarm cannot escape from a local optimum~\cite{nezami2013dynamic}. To address this, a hybrid PSO with Gaussian mutation is proposed to promote diversity to avoid early convergence. Then, K-means is applied to the centroids obtained by PSO to refine their location and get the best possible solution. Our methodology performs activity discovery using unsegmented input data and the proposed techniques used are unsupervised with no prior knowledge of the labels of the different activities.
The main contributions of this paper are: 
\begin{itemize}
\item A methodology consists of keyframe selection, feature extraction, and PCA to represent human activities. The features based on displacement, statistical and orientation are extracted simultaneously to represent all the movement aspects of human activities. This makes the discovery part perform better because it has comprehensive information about the activities.
  \item A hybrid clustering algorithm called HPGMK to discover and group unlabelled human activities observations into individual activity classes. PSO is customized by applying the Gaussian mutation, based on the advantages of two methods~\cite{jana2019repository} and ~\cite{li2008adaptive}, on the global best's centroids to increase the diversity of selected clusters in the global best. Also, due to the increase in the variety of solutions, the proposed algorithm reaches the desired solution in a smaller number of iterations. That is why the clustering time is reduced.
  \item Integrating K-means to refine the obtained cluster centers from the PSO to improve the exploitation of the algorithm. In PSO, when approaching the final solution, the speed of the particles decreases greatly and it becomes difficult to reach the best optimum solution. For this reason, the problem of PSO is solved by using the advantage of k-means in local search and applying it to the obtained solution from PSO.
  \item An evaluation of the performance of the proposed methodology compared with the latest fully unsupervised methodology for HAD and the state-of-the-art HAD and deep clustering methods. The proposed method demonstrates superior performance on five challenging 3D skeleton-based activity analysis datasets. 
\end{itemize}
In this paper, the background and related methods are discussed in section~\ref{Related}. The methodology is described in section~\ref{Proposed}. We present the evaluation of the proposed approach, comparing with state-of-the-art (SOTA) techniques in Section~\ref{Experiments}, and finally, the conclusion is stated in Section~\ref{Conclusion}. 

\section{Related works}\label{Related}

\textbf{Feature extraction from 3D skeletal human activities data.}
Skeletal data includes the number of joints, and each joint contains three-dimensional coordinates. Since the motion of the joints has essential information for any activity, feature extraction is vital. There are various methods for representation of the motion of skeletal joints such as calculating the difference between the joints in the same frame and the same joints in different frames~\cite{adama2018human}, using Histogram Oriented of Joints~\cite{xia2012view}, Dynamic Time Warping algorithm~\cite{tabejamaat2022embedded}, Covariance of 3D Joints~\cite{hussein2013human}, generating joint rotation matrix concerning the person's torso~\cite{sung2012unstructured}, and extracting the angles and orientations of the most informative body joints~\cite{cao2019geometric}. However, most of these methods extract one aspect of the skeletal data features, leading to other important aspects of activities being overlooked. As a result, there is a decrease in accuracy in the final result because of the insufficient discriminating ability of the extracted features. Moreover, due to the complexity of feature calculations, some of these methods cause computational latency. The difference between our paper with previous works for extracting features is that We have combined the feature extraction techniques from \cite{adama2018human,liu2020rotation,cao2019geometric} to extract three skeletal features from informative joints and keyframes. Previously, these features have been applied separately and most of them have used all of the joints and frames, which increases additional information. This increases the time complexity and reduces identifying the activities' performance. Some methods like \cite{shan20143d} and \cite{arzani2020switching} have tried to select some frames that are more distinguishing compared to other frames and remove redundant information with the assistance of kinetic energy. However, these methods are applied to each activity sample separately. In other words, in \cite{shan20143d} and \cite{arzani2020switching},  keyframes were selected in a supervised manner. In contrast, we apply the above methods in our proposed method to select keyframes without knowing the beginning and end of the activities.\\
\textbf{Particle Swarm Optimization Clustering and dimension reduction.}
One of the important methods to do discovery is to use clustering. Clustering is a method that categorizes data points based on similarities and dissimilarities. One of the common methods of clustering is K-means. But it has problems such as poor convergence rate and local optimum. One of the ways to overcome these problems is to use evolutionary algorithms like PSO. PSO is a population-based algorithm where each individual represents a potential solution, which makes a better problem space search. PSO greatly reduces the possibility of getting stuck in local optimum due to using local search and global search simultaneously. To create a novel clustering algorithm, Malarvizhi et al.~\cite{malarvizhi2021data} combined PSO and feature linkage-based weight reduction. The method determined the weight of each feature using the Mahalanobis distance to choose the feature and do the clustering automatically. Sharma et al.~\cite{sharma2019sustainable} developed a hybrid PSO clustering for network-based sustainable computing. They applied the mutation operator to ensure diversity among the solutions to keep the algorithm's balance between exploration and exploitation. Rengasamy et al.~\cite{rengasamy2021pso} Introduced a new memory dimension termed family memory and added to the two already existing ones of cognitive memory and social memory. This memory was used to collect the information from the particles that favor a certain cluster. Additionally, they utilized the K-means to initialize centroids for PSO clustering to enhance the traditional PSO. Cai et al.~\cite{cai2020novel} proposed a new clustering method based on combining density peaks clustering with PSO. They employed a technique to compute density peaks to avoid falling into a local. They also presented a new fitness criteria function to optimally explore K cluster centers to obtain the optimal global solutions. However, these methods do not address all the weak points of PSO. Some of them either deal with the issue of reducing the speed of PSO when approaching the optimal solution or the issue of reducing the diversity among the solutions during the search. Different from the above-mentioned methods, we overcome the weakness noted for PSO simultaneously by using Gaussian mutation and K-means. By employing Gaussian mutation, the variety of the solutions the PSO's capacity to exploit around potential solutions gets effectively enhanced. We also utilize K-means to solve the PSO problem: its convergence speed slows down when it approaches the global optimum.
We also use K-means to solve the problem of reducing the convergence speed of PSO when approaching the global optimum by using the fast speed of K-means in local search. Zhang et al.~\cite{zhang2021clustering} introduced a feature selection method based on PSO that combined fuzzy clustering and feature importance (PSOFS-FC). They presented a new objective function based on F-measure and filling risk for PSO with fuzzy clustering to assess the impact of missing data in class imbalance.
To overcome the dimensionality curse, Song et al.~\cite{song2021fast} presented a three-phase feature selection technique based on correlation-guided clustering and particle swarm optimization. First, they combined a filter approach and a feature clustering-based method to reduce the search space. Then, an enhanced integer PSO was used to select the best feature subset.
Unlike PSOFS-FC and other mentioned methods, which combine clustering and feature selection, we reduce extracted features' dimensions before clustering. Although good results have been obtained in these methods, due to the simultaneity of feature selection and clustering, clustering becomes problematic in high-dimensional data and the clustering time increases greatly. In HAD, spending time to discover activities is important because of its applications, such as use in security areas to identify suspicious behavior or in hospitals to check patients' status. For this reason, our proposed method reduces the feature dimensions before clustering by using PCA to make the data more clusterable. PCA speeds up the clustering algorithm by removing correlated features that do not influence decision-making. As a result, the algorithm's clustering time decreases dramatically with fewer features. Thus, not only the speed of clustering increases but also the accuracy of clustering increases because the clustering process is performed on highly important features.
Regarding detecting outlier and noisy data, Hubert et al.~\cite{hubert2004robust} proposed a combination method to make PCA robust to outliers. They combined projection pursuit~\cite{hubert2002fast} with h robust covariance estimation in lower dimensions~\cite{hubert2005robpca}. Moreover, they applied a diagnostic plot to detect the outliers. Candes et al.~\cite{candes2011robust} proposed a technique to improve the performance of PCA. They used a low-rank and sparse component for PCA to avoid outliers and achieved good performance in the application of Alzheimer's Disease Recognition~\cite{alessandrini2022eeg}. Rahmani et al.~\cite{rahmani2019outlier} presented a provable algorithm to identify the outliers based on PCA. For this reason, they employed a convex optimization problem to evaluate the data points based on the innovation search method. Despite the very good performance of the presented methods to improve PCA, they have more execution time than the original PCA. On the other hand, the focus of our work is on the improvement of feature extraction and discovery for human activities. As mentioned before, time is very important in HAD. That is why we use PCA to prevent the increase in the computation of time for the presented framework. It is worth mentioning that the obtained results show that PCA has reduced the dimensions of the features and improved HAD performance significantly.
\textbf{Recognition and discovery of 3D skeletal human activity.}
Many studies in HAR used supervised approaches~\cite{shi2019two,li2021memory,gaglio2014human}. Yadav et al.~\cite{yadav2022skeleton} combined long-short term memory networks and convolutional neural networks for recognizing human activity and fall detection. They used some handcrafted features, including geometrical and kinematic features to guide their proposed model. Zhang et al.~\cite{zhang2020semantics} proposed an end-to-end semantics-guided neural networks framework. They provided two semantic forms based on joints and frames and used GCN and CNN layers to find the dependence of joints and frames, respectively. Si et al.~\cite{si2019attention} proposed a novel model based on a recurrent network. They applied a graph convolutional layer into the LSTM network to improve the performance of traditional LSTM. They also introduced an attention gate inside the LSTM to capture discriminative features. Xia et al.~\cite{xia2021multi} provided a graph convention network based on spatial and temporal. They applied an attention layer to the model to generate discriminative features and modified feature maps. Then, a softmax classifier was used to categorize the activities. Cai et al.~\cite{cai2021jolo} introduced a scheme to capture visual information surrounding each skeleton joint and achieve local motion cues. They extracted features from skeleton and RGB data using two graph convolutional networks. Then, both types of features were concatenated and activities were classified by calculating a score based on linear blending. The problem with these approaches is that they require activity labels in the training data. Humans annotated the labels during data preparation. It makes these methods impractical with real-life data that are mainly unlabeled. Our work does not use labels for training in our algorithm. The algorithm discovers activities by looking for similar features between them. In addition, the methods mentioned above use deep learning techniques. In contrast, as a first study, we do not use them in our method and the focus is on developing a comprehensive model for HAD as a baseline. \\
Several approaches try to address the HAR in an unsupervised way. Wang et al.~\cite{wang2021deep} presented a deep clustering method based on a dual-stack auto-encoder to map raw data to spatio-temporal features. After extracting features, the radial basis function neural network was used to classify the activities. Su et al.~\cite{su2020predict} provided an unsupervised model by employing a bi-directional recurrent neural network and used K-NN to classify the activities. Liu et al.~\cite{liu2022spatial} designed a spatial-temporal asynchronous normalization method to reduce redundant information related to time and normalize the spatial features. Next, they used a gated recurrent unit auto-encoder to feature vectors. First, all of these methods received the activities already segmented which has enabled them to be aware of the differences between the activities before performing the recognition. Second, in most of these methods, only feature extraction was performed without supervision. The supervised classification method was used for the rest of the operations to learn activity models using activity labels.\\
On the other hand, human activity discovery can automatically categorize human activity in a fully unsupervised way and the challenge is learning from unlabeled data. The majority of existing methods were developed for sensor-based~\cite{gupta2020tracking,qi2020smartphone} and RGB video data ~\cite{yan2020synergetic,zhang2015fuzzy}.~The challenges of the sensor-based approach are difficult to implement in the environment and take a long time to install ~\cite{ong2015autonomous}. Furthermore, it is impractical for people to wear sensors everywhere. With RGB videos, the problems faced are millions of pixel values, illumination variations, viewpoint changes, and cluttered backgrounds ~\cite{han2013enhanced}. In this work, we concentrate on 3D skeleton-based data as it does not have the problems of the other two data types. One of the first works in HAD was performed by Ong et al.~\cite{ong2015autonomous}. They proposed an autonomous learning technique based on the mixture of the Gaussian hidden Markov model. They introduced an incremental clustering approach based on k-means to discover the activities to deal with the undefined number of clusters. An issue with their approach is that they have used k-means to discover the activities that get stuck into the local optimum easily and they have not examined all aspects of the skeleton data features. Moreover, they extracted all the features from all joints, resulting in more redundant data and increased discovery errors.
Recently several approaches have been proposed by~\cite{paoletti2021subspace} to solve HAR without labels. In their proposed methods, several clustering methods, including spectral clustering (SC), elastic net subspace clustering (ENSC), and sparse subspace clustering (SSC) were used, which used covariance descriptors. They used an affinity matrix to find similarities and then applied spectral clustering. In addition, a time stamp pruning approach was used to remove redundant data to normalize temporal features. Although they have achieved impressive results, the data used were already segmented by activity before performing discovery. It means that the activities are already categorized. Because each sample contains an activity that performs completely. In other words, the beginning and end of each activity are clear.\\
In a nutshell, many methods have been proposed to recognize human activities in a supervised and unsupervised manner and have obtained very acceptable results. But the problem with these methods is that they ignore the discovery step. These methods are useless because labeling activities do not occur in real-time. They also need a lot of computation time for training. In reality, training data are not available. If we have a dynamic big and growing video related to human activity, we are not sure of the labels to predefine the rules. This can be a real challenge. On the other hand, due to the variety of human activities, these methods need to be retrained for new activities, making them not scalable.
In the case of HAD, in addition to the fact that there are very limited methods, these methods have problems including using shallow methods which are not very accurate or not fully performing the discovery process in an unsupervised manner.
In this paper, we propose an approach to discovering activities from untrimmed videos without knowing the label of activities. It makes this method suitable for use in real-world scenarios. In addition, we use a feature extraction approach to examine most aspects of skeleton data along with a keyframe selector to reduce redundant information and discovery accuracy.
\section{Proposed human activity discovery}\label{Proposed}

\begin{figure*}
 \begin{center}
  \includegraphics[height=5.5 in, width=\textwidth]{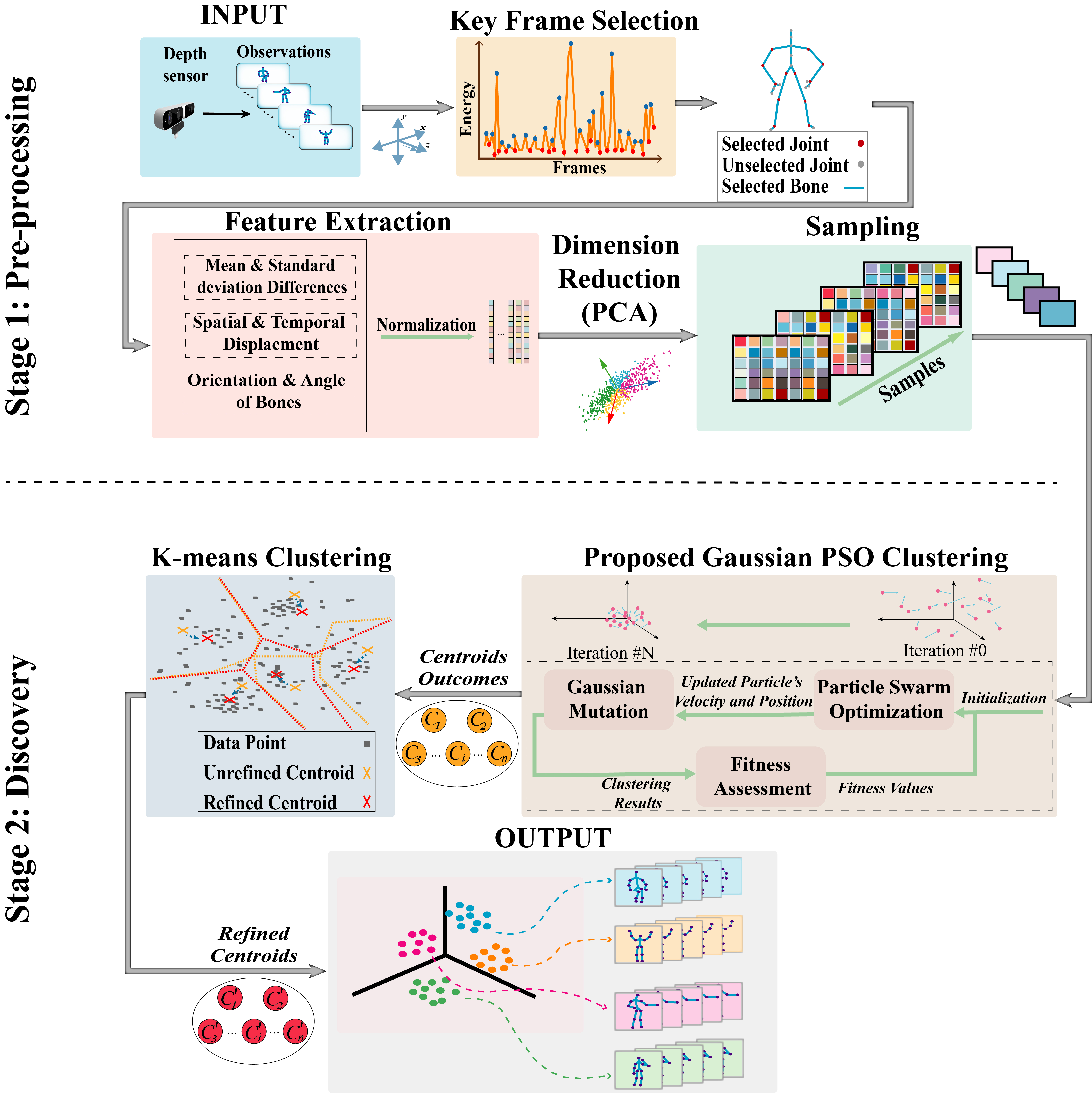}
  \caption{Methodology of the proposed approach has two stage pre-processing and discovery. In pre-processing, keyframes are selected from the video sequence by computing kinetic energy.  Then, features based on different aspects of skeleton including displacement, orientation, and statistical are extracted from informative joints and bones. Principal components are then chosen by applying PCA on the features. Next, overlapping time windows is used to segment a series of keyframes as activity samples. In discovery stage (HPGMK), Hybrid PSO clustering with Gaussian mutation operator is used to discover the groups of activities. Eventually, K-means clustering is applied to the resultant cluster centers to refine the centroids.}\label{proposed method}
  \end{center}
\end{figure*}

The proposed framework consists of two main stages. In the first phase, we propose an approach that can extract high-quality features in an unsupervised manner. Three crucial factors should be taken into account in this matter. First, not all the captured frames are important. Due to the similarity of frames and noises, HAD performance reduces sharply. Therefore, we need frames that show the salient features of the activities. Also, not all the Joints have an effective role in discovering activities, such as the torso, which is constant in most activities. Extracting features from these joints increase time complexity. Second, the extracting features should accurately reflect all aspects of human activity by considering all factors such as both spatial and temporal. Third, the feature dimension needs to be minimal to make the clustering operation easier (a larger dimension confuses the clustering process). To cover the first phase, we present a pre-processing method as shown in Fig.~\ref{proposed method}(Stage 1). We employ an innovative approach based on kinetic energy to select representative frames of video sequences as keyframes. In other words, we seek to select the frames that show the most prominent characteristics of the activities as the keyframe. However, the selection of particular Keyframes without losing the required information is a challenging task. When only the local maximum kinetic energies are considered keyframes, the sequence of activities may break up and no longer represent activities. For example, in the walking activity, considering keyframes with high local energy, only positions where both legs move away from each other are considered keyframes. In contrast, positions where both legs are placed together show a part of the walking process that is lost in this state. For this reason, we employ keyframe selection based on local maximum and minimum kinetic energy that can find representative frames and reduce complexity. Moreover, it can maintain the order of the activity. To avoid increasing computational time and overlapping among activities, we select joints (informative) that have a vital role in displaying activities based on experimental tests. To increase the discovery performance, it is necessary to extract features to represent all aspects of each activity. for this purpose, we design a method to represent the activities based on spatial and temporal displacement, statistical, and orientation features. The displacement-based representations provide the view-invariant Spatio-temporal human representations due to invariant to positions and orientations of humans with respect to the camera. To obtain features that are invariant to human scale changes, orientation-based representations are extracted to find relative information between human joints. Statistical features describe how activity evolves over time, especially when separating actions involving the arms and legs. Therefore, statistical time-domain characteristics represent changes in a set of postures for a time-domain activity. We employ a PCA to address the third factor to reduce dimensionality and make the extracted feature more clusterable while high-importance features are kept. We adopt a sliding window over the untrimmed skeleton streaming sequences to perform activity discovery. To increase the number of samples and avoid pruning important events like a transition between activities we employ overlapping sliding windows increases performance. In the second phase, we propose a novel clustering algorithm based on hybrid PSO called HPGMK to discover human activities (Fig.~\ref{proposed method}(Stage 2)). The key benefit of PSO is that there are fewer parameters to set. Contrary to genetic algorithms, PSO does not use complex evolutionary operators like the crossover, making it less complicated. However, the issue is that its convergence speed slows down when it approaches the global optimum. For this reason, we combine PSO with k-means to use the fast speed of k-means to reach the local optimum to improve PSO's performance. The Gaussian mutation is also employed on the global best particle to search for areas around it that have a high potential to be selected and generate diverse solutions and strike a balance between exploitation and exploration. In this section, the proposed method is presented, as shown in Fig.~\ref{proposed method}. Keyframes are first extracted from the input video based on the kinetic energy of all frames. Three types of features including spatial and temporal displacement, mean and standard deviation differences, and orientation and angle features are extracted. Next, feature vectors are reduced by PCA and frames are sampled in specific periods to segment the activity steam. Finally, the proposed clustering technique assigns each sample to an appropriate category of activities. The details of each part of the methodology are described below.
\subsection{Keyframe selection}
\begin{figure}
  \begin{center}
  \includegraphics[width=\textwidth,keepaspectratio]{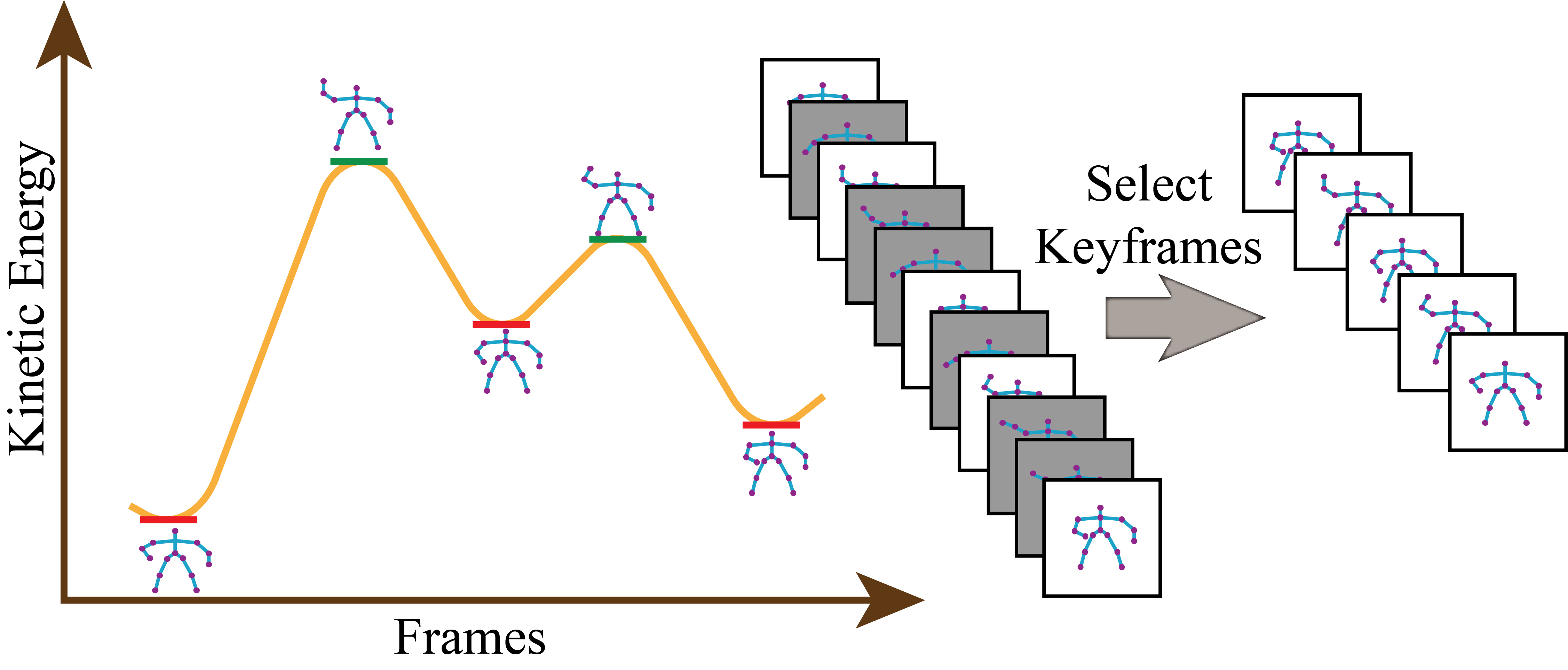}\\
  \caption{Illustration of the keyframe selection.}\label{Keyframe}
  \end{center}
\end{figure}

Keyframe selection is a process of selecting frames reflecting the main activities in the video. Some methods like \cite{shan20143d} and \cite{arzani2020switching} have tried to select some frames that are more distinguishing compared to other frames. However, these methods were applied to each activity sample separately. In other words, in methods \cite{shan20143d} and \cite{arzani2020switching},  keyframes were selected in a supervised manner. In our proposed method, keyframes are selected without knowing the beginning and end of the activities. To find the keyframes, the kinetic energy \textit{E($f_i$)} of each frame \textit{$f_i$} is calculated ~\cite{shan20143d} using Eq.(\ref{keyframe}), based on the displacement of joints over time. In this way, the movement of a joint \textit{j} between frame \textit{i} and \textit{i}-1 is calculated for all joints (\textit{J}). The sum of the movements for all joints is the energy of the current frame. Frames with local maxima and minima amount of kinetic energy compared to neighboring frames are considered keyframes (see Fig.~\ref{Keyframe}) Because these are the energy's extreme points, which are meant to resemble crucial posture data.
\begin{equation}\label{keyframe}
    E(f_{i}) = \sum_{j=1}^{J}E(f^{j}_{i})=1/2 \sum_{j=1}^{J}(f^{j}_{i}-f^{j}_{i-1})^2
\end{equation}
\subsection{Feature extraction}
 To represent the activities, a set of statistical displacements, angles and orientation features for encoding key aspects of activities are extracted. These important features are extracted from selected (informative) joints in the data to describe the shape and movement of human. Selected joints have been obtained based on experimental tests that included left and right hand, foot, hip, shoulder, elbow and knee. These joints have more movement and contribution than other joints such as torso in activities. We use information related to the position and movement of joints, the orientation and angle between a pair of bones and activity variation over time. The normalization procedure ~\cite{adama2018human} is performed on all features.
\subsubsection{Displacement features}
Joint displacement-based features encode information on the position and motion of body joints, particularly displacement between joints of a pose and position differences of skeleton joints across time~\cite{adama2018human}. 
\begin{itemize}
\item Spatial joint displacement is computed using pairwise Euclidean distances between joints \textit{$P_{i}$} and \textit{$P_{j}$}  (\textit{i} $\neq$ \textit{j}) in 3D space in the same frame, Eq.(\ref{Spatial}). The joint pairs used are both hands, hands and head, and hip and feet at both sides, giving 5 features per frame, (see Fig.~\ref{spatial and temporal displacement}(a)).
\begin{equation}\label{Spatial}
    Pairwise Distances=\sqrt{\sum_{x,y,z}(P_{i}-P_{j})^{2}}
\end{equation}
\item Temporal joint displacement is calculated based on two modes. \textit{$T_{cp}$} is the difference between each selected joint \textit{$P_{i}$} in current frame \textit{$P^{c}_{i}$} and previous frame \textit{$P^{c-1}_{i}$} (see Fig.~\ref{spatial and temporal displacement}(b)) to determine the small changes in joint movement over time (Eq.(\ref{Temporal})). \textit{$T_{cn}$} is the difference between each selected joint of current frame and the frame of neutral pose (We randomly select a standing position as a neutral position) \textit{$P^{n}_{i}$}, illustrated in Fig.~\ref{spatial and temporal displacement}(c), to find general changes in joint movements as given in Eq.(\ref{Temporal_netural}).
\begin{equation}\label{Temporal}
    T_{cp} = P^{c}_{i}-P^{c-1}_{i}
\end{equation}
\begin{equation}\label{Temporal_netural}
    T_{cn} = P^{c}_{i}-P^{n}_{i}
\end{equation}	
\begin{figure}
  \begin{center}
  \includegraphics[width=\textwidth,keepaspectratio]{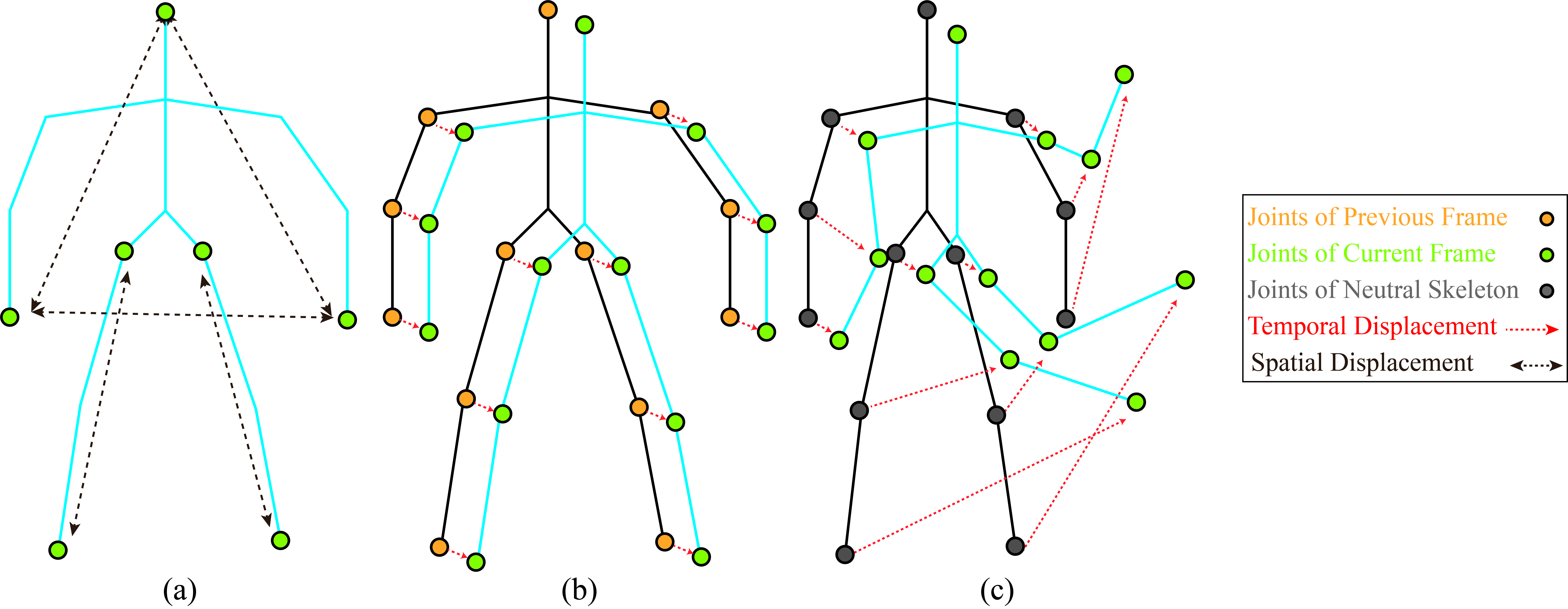}\\
  \caption{(a) Spatial displacement of pairwise joints in the same frame. Temporal displacement of current frame from (b) previous frame and (c) the neutral frame.}\label{spatial and temporal displacement}
  \end{center}
\end{figure}
\end{itemize}
\subsubsection{Statistical features}
The mean and standard deviation of time-domain features express how activity changes over time, particularly in distinguishing between activities related to the arms and legs. Thus, statistical time-domain features encode variations across a collection of poses of an activity in time-domain. These features are calculated by the difference of selected joint \textit{$P_{i}^c$} in current frame from mean \textit{$P_{(i,mean)}$} and standard deviation \textit{$P_{(i,std)}$} of the selected joint coordinates within an activity sequence as given by Eq.(\ref{mean difference}) and (\ref{standard deviatio_main})~\cite{adama2018human}.
\begin{itemize}
\item Joint coordinate-mean difference
\begin{equation}\label{mean difference}
    P_{i_{(mean)}}^c = P_{i}^c-P_{(i,mean)}, P_{(i,mean)}=\frac{1}{N}\sum_{c=1}^{N}P_{i}^c
\end{equation}
N is the number of frames.
\item Joint coordinate-standard deviation difference
\begin{equation}\label{standard deviatio_main}
    P_{(i,std)}^c = P_{i}^c-\sqrt{\frac{\sum_{i=c}^{N}(P_{i}^c-P_{(i,mean)})^{2}}{N}} 
\end{equation}
\end{itemize}
\subsubsection{Orientation features}
The three-dimensional coordinate system \{\textit{x},\textit{y},\textit{z} $\in$ \textit{$R^3$}\} represents points as joints. \textit{x}, \textit{y}, and \textit{z} denote the 3D coordinates of joints. Joints and bones can be described by the orthonormal vectors ~\cite{liu2020rotation} as follow:
\begin{equation}\label{standard deviation}
    P^{f}_{i} = x^{f}_{i}e_{1}+y^{f}_{i}e_{2}+z^{f}_{i}e_{3}
\end{equation}
\begin{equation}\label{standard deviation}
    B^{f}_{ij} = (x^{f}_{i}-x^{f}_{j})e_{1}+(y^{f}_{i}-y^{f}_{j})e_{2}+(z^{f}_{i}-z^{f}_{j})e_{3}
\end{equation}
where $P_i^f$ is the \textit{i}th skeleton joint in the \textit{f}th frame and {\textit{$e_1,e_2,e_3$} } are orthonormal vectors. \textit{$B_{ij}^f$} is the bone between two adjacent joints \textit{$P_i^f$} and \textit{$P_j^f$}. Moreover, magnitude and direction of two bones \textit{a} and \textit{b} are represented by geometric product where this product is the sum of internal ($a . b$) and external ($a\wedge b$) product.
where the inner product is used to compute the length and angle between two bones \textit{a} and \textit{b}. The outer product of two bones can be regarded as an oriented plane containing  \textit{a} and  \textit{b}.
The orientation and angles between bones features are obtained in the process described as follows. 
\begin{itemize}
\item The rotation matrix is a transformation matrix that describes the rotation from a bone to another. Three angles are required to define the rotation matrix between two bones. The rotation angles are considered as orientation features.  The elements of these features are the rotation of bones relative to the x, y, and z axes (see Fig.~\ref{oriantation}(a)). 
\item The angle features consist of the angles between the bones of elbow-wrist and shoulder-elbow at both sides and the angles between the bones of hip-knee and knee-ankle at both sides. These angles are highlighted in Fig.~\ref{oriantation}(b) and calculated as below:
\begin{equation}\label{angle}
    \theta = 180 \times \frac{arctan^{2}(\frac{\norm{bone_{i}\wedge bone_{j}}}{\norm{bone_{i} . bone_{j}}})}{\pi } +180
\end{equation}
where \textit{$bone_{i}$} and \textit{$bone_{j}$} are determined by Eq.(\ref{standard deviation})
\begin{figure}
  \begin{center}
  \includegraphics[width=\textwidth,keepaspectratio]{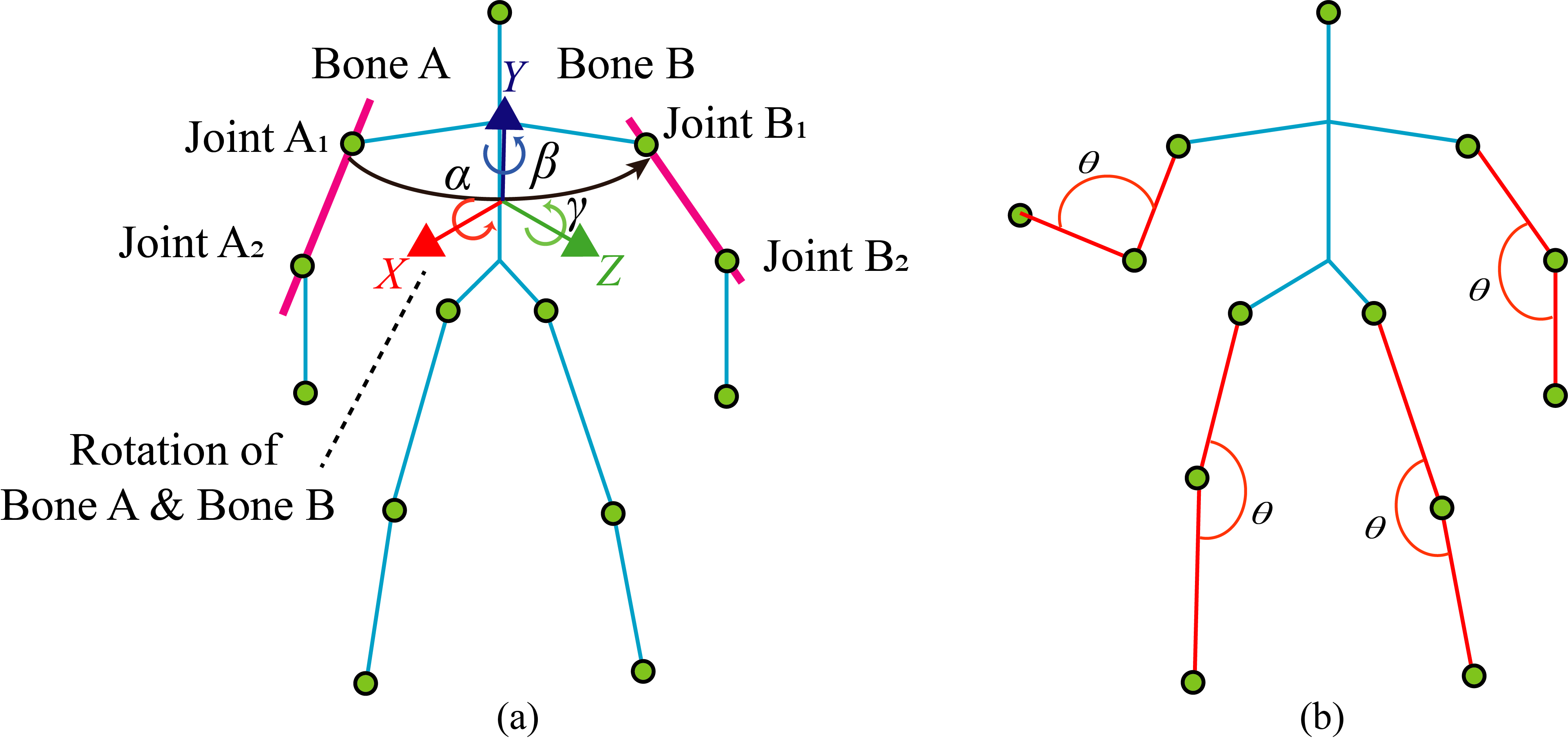}\\
  \caption{(a) Illustration of the rotation between two bones A and B. $\alpha$, $\beta$, and $\gamma$ are the orientation of angles. (b) The angles of the selected body bones. The angles of elbow-wrist and shoulder-elbow at both sides and angles between the bones of hip-knee and knee-ankle at both sides are used to calculate angle features.}\label{oriantation}
  \end{center}
\end{figure}
\end{itemize}
\subsection{Feature selection and sampling}
\begin{figure}
  \begin{center}
  \includegraphics[width=\textwidth]{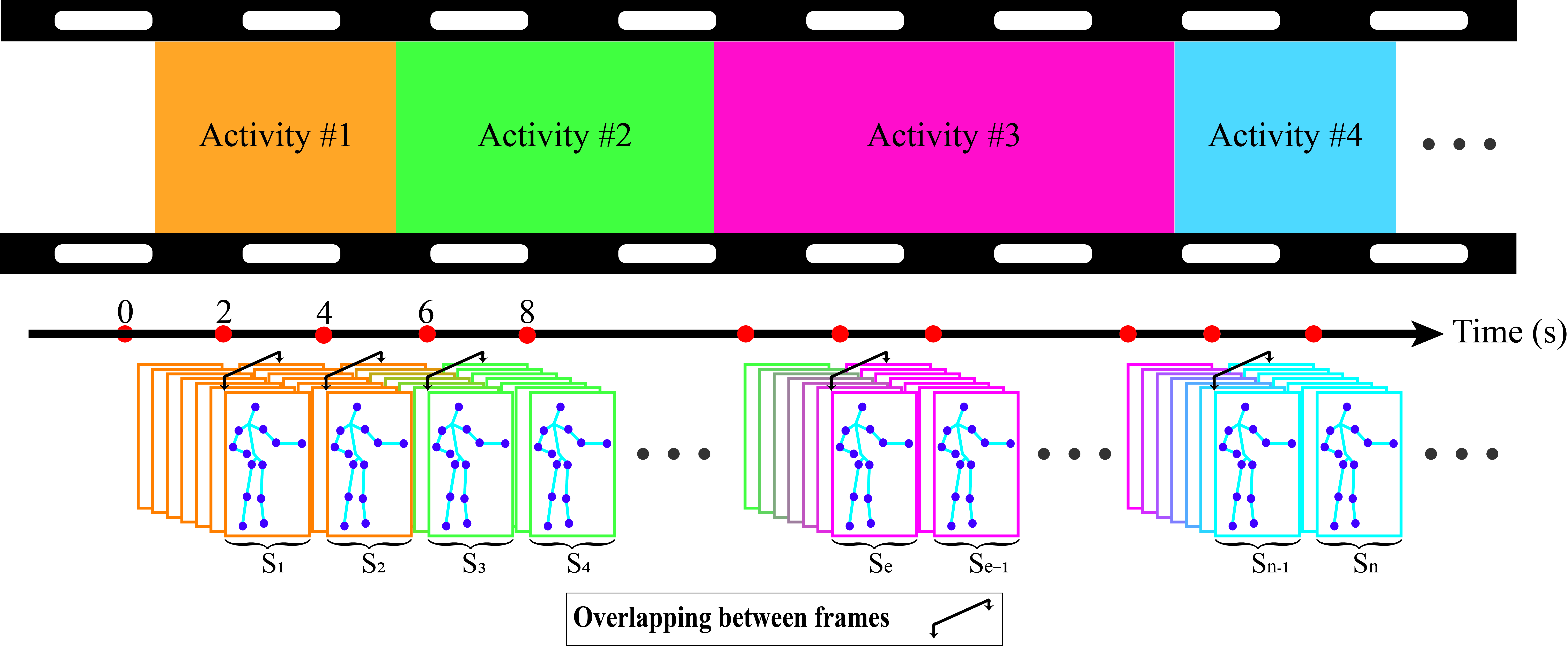}\\
  \caption{Illustration of sampling based on overlapping sliding windows. Each sample ($S_1$, $S_2$,$\ldots$, $S_n$), except the first sample, starts with the last frame of the previous sample.}\label{Sampling}
  \end{center}
\end{figure}

For fast clustering and complexity reduction, key features are extracted by PCA. Then, sliding windows are used to segment frames into time windows. Each window comprises of 15 frames. The overlap of sliding windows increases performance. Because it increases the number of samples and avoid pruning important events like transition between activities~\cite{presti20163d}. The first 15 frames do not overlap while in other samples, their first frame starts from the last frame of the previous sample (see Fig.~\ref{Sampling}).

\subsection{Proposed clustering}
\begin{figure}
  \begin{center}
  \includegraphics[width=\textwidth]{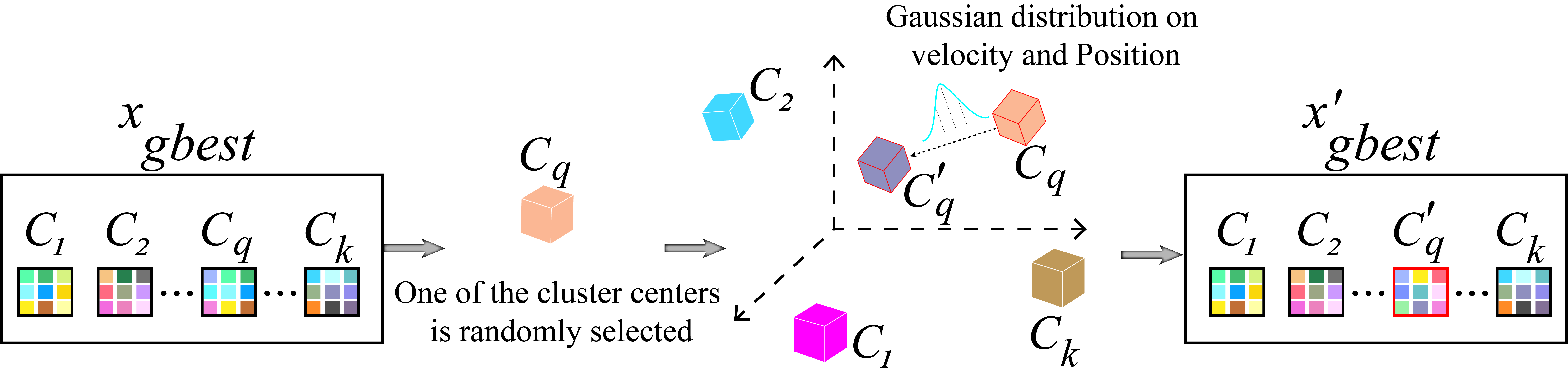}\\
  \caption{Visualization of the Gaussian Mutation Operator. In each iteration of hybrid PSO, one centroid ($C_{q}$) is chosen from $x_{gbest}$ randomly. Then, Gaussian distribution is applied on position and velocity of the selected centroid based on Eq.(\ref{velocity Gaussian mutation}) and (\ref{pos Gaussian mutation}) to create a new offspring $C_{q}^{'}$. The new global best ($x^{'}_{gbest}$) is then compared to $x_{gbest}$. If $x^{'}_{gbest}$ has better fitness value than $x_{gbest}$, $x^{'}_{gbest}$ is replaced with new global best.}\label{Gaussian}
  \end{center}
\end{figure}

PSO is a population-based optimization algorithm ~\cite{hadikhani2020adaptive}. A population is made up of a number of particles and each particle represents a solution and moves according to its speed. The changes in velocity and position of the particles are calculated based on the following formula: 
\begin{equation}\label{velocity}
    x_{i}(t+1)= x_{i}(t)+v_{i}(t) 
\end{equation}
\begin{equation}\label{position}
\begin{split}
    v_{i}(t+1)=w\times v_{i}(t)+c_{1}\times rand_{1}\times (pbest_{i}(t)-x_{i}(t))\\
    +c_{2}\times rand_{2}\times (gbest(t)-x_{i}(t))
\end{split}
\end{equation}
\begin{equation}\label{weight}
    w=\frac{w_{max}+t\times (w_{max}-w_{min})}{t_{max}}
\end{equation}
\begin{equation}\label{c1}
    c_{1}(t+1)=(c_{1_{max}} - c_{1_{min}})\times \frac{t}{t_{max}}+c_{1_{max}}
\end{equation}
\begin{equation}\label{c2}
    c_{2}(t+1)=(c_{2_{max}} - c_{2_{min}})\times \frac{t}{t_{max}}+c_{2_{max}}
\end{equation}
In Eq.(\ref{velocity}) and (\ref{position}) \textit{$x_{i}$}(\textit{t})  and \textit{$v_i$}(\textit{t}) are the position and velocity of the particle \textit{i} at time \textit{t} respectively. \textit{$pbest_{i}$} is the best position found by the particle \textit{i}. \textit{$gbest$} is the best position found in the population. $w$ is the inertial weight defined by Eq.(\ref{weight}) and starts to decrease from \textit{$w_{max}$}. \textit{$c_{1}$} and \textit{$c_2$} are acceleration coefficients expressed by Eq.(\ref{c1}) and (\ref{c2}). The \textit{$c_{1_{max}}$}, \textit{$c_{2_{max}}$} and \textit{$c_{1_{min}}$}, \textit{$c_{2_{min}}$} are initial and final values, respectively, \textit{t} is the number of iterations and \textit{$t_{max}$} is the maximum number of iterations~\cite{cai2020novel}. \textit{$rand_1$} and \textit{$rand_2$} are random variables between 0 and 1.  Each solution is evaluated by Eq.(\ref{SSE}) which should be minimized to achieve proper clustering. 
\begin{equation}\label{SSE}
    SSE = \sum_{k=1}^{K}\sum_{\forall x\in c_k} \|x_i-\mu_k\|^2 
\end{equation}
\textit{$x_i$} is a data point belonging to the cluster \textit{$C_k$} and \textit{$\mu_k$} is the mean of the cluster \textit{$C_k$}. $k$ is the number of clusters specified. To avoid in local optimum, a Gaussian mutation operator based on ~\cite{jana2019repository} and ~\cite{li2008adaptive} is applied to the global particle as follows:
\begin{equation}\label{velocity Gaussian mutation}
\begin{split}
   v^{'}_{gbest}(d)=v_{gbest}(d)\times  G(0,h)\\ \times (x_{max}(d) - x_{min}(d))
\end{split}
\end{equation}

\begin{equation}\label{pos Gaussian mutation}
    x^{'}_{gbest}(d)=x_{gbest}(d)+G(0,h)\times v^{'}_{gbest}(d)
\end{equation}
where $x_{gbest}$ and $v_{gbest}$ represent the position and velocity of global best particle. $x_{max}$ and $x_{min}$ are the maximum and minimum value in $d^{th}$ dimension. Gaussian (0,\textit{h}) is Gaussian distribution with the mean 0 and the variance \textit{h}.
the value of \textit{h} is start with a high value to increase the exploration ability of the algorithm to find interested region at the beginning of the search. Then \textit{h} decreases linearly during each iteration according to Eq.(\ref{variance}), to increase the power of exploitation at the end of search to reach optimum solution. 

\begin{equation}\label{variance}
    h(t+1)=h(t)-(1/t_{max})
\end{equation}
where \textit{$t_{max}$} is the maximum number of iterations. Fig.~\ref{Gaussian} is an illustration of the Gaussian mutation.\\
In general, the core of HPGMK is based on PSO which is a population-based algorithm. The position of each particle in the population represents a solution. In other words, each particle contains the position of the cluster centers. Each particle updates its position using its velocity to reach the optimum solution~\cite{kennedy1995particleadssd}. Furthermore, we have one objective function which is SSE (sum square error). We evaluate each individual based on SSE (Eq.(\ref{SSE})). In this process, an individual is chosen as the global best in each iteration with the lowest SSE value among the rest of the individuals. To increase the diversity of solutions, a Gaussian mutation is applied to the position and velocity of the global best particle.

The velocity of the particles reduces quickly as PSO approaches the global optimum, and in most circumstances, the ideal solution is not achieved. For this reason, K-means is applied to the obtained centroids from PSO to refine them.~After the completion of the PSO process, the global best solution is selected based on the SSE value. Then, the selected solution is modified by averaging the position of data points in each cluster to select the best position for the cluster centers. This continues until the position of the clusters does not change.
The routine of the proposed clustering algorithm is shown in Algorithm ~\ref{HPGMK-algo}.

\begin{algorithm}[!htbp]

\DontPrintSemicolon
  
  \KwInput{D=\{$d_1$,$d_2$,$\ldots$,$d_n$\} //Set of data points\;~~~~~~~~~~~~$k$ //Number of desired activities (clusters)}
  \KwOutput{ Set of $k$ clusters}
  Initialize a population of particles with random positions and velocities in the search space\;
  \For{t=1 to the maximum number of iteration}
  {
    \For{each particle i }
    {
        Update position and velocity of particle \textit{i} according to Eq.(\ref{velocity}) and Eq.(\ref{position})\;
        Evaluate fitness value of particle \textit{i} according to the fitness function in Eq.(\ref{SSE})\;
        Update $pbest_i$(t) and \textit{gbest}(t) if necessary\;
        \For{T times (T is the number of iteration
for mutation and set 10)}
        {
            Mutate \textit{gbest}(t) according to Eq.(\ref{velocity Gaussian mutation}) and (\ref{pos Gaussian mutation})\;
            Compare mutated \textit{gbest}(t) with previous and choose the best as new \textit{gbest}(t)\;
        }
    }
  
  }
  Use \textit{gbest}(t) as the initial centroids\;
    
   \While{until no change}
   {
   \tcp{Refining the centroids}
        Calculate distances of data points to centroids\;
        Assign data points to the closest cluster\;
        Centroids are updated using the following equation
        $centroid_i= \frac{1}{n_i}\sum_{\forall d_i \in C_i} d_i$,~~\\
        where $n_i$ is the number of data points in the cluster \textit{i}
   		
   }

\caption{Hybrid PSO with Gaussian Mutation and K-means (HPGMK)}
\label{HPGMK-algo}
\end{algorithm}

\section{Experiments}\label{Experiments}
\subsection{Datasets}
Five datasets were used to evaluate the effectiveness of proposed method: Cornell Activity Dataset (CAD-60)~\cite{sung2012unstructured}, UTKinect-Action3D (UTK)~\cite{xia2012view}, Florence3D (F3D)~\cite{seidenari2013recognizing}, Kinect Activity Recognition Dataset (KARD)~\cite{gaglio2014human}, and MSR DailyActivity3D (MSR)~\cite{wang2012mining}. These datasets have different dimensions, features, and activities. Table \ref{tabel1} shows the statistical information of these datasets. They are discussed as follows.
\begin{table}[h!]
\caption{Number of activities, subjects and  videos in the five datasets used}
    \centering

    \begin{tabular}{c|c|c|c|c|c}
             \hline
    \textbf{Dataset}&\textbf{CAD-60}& \textbf{UTK}& \textbf{F3D}& \textbf{KARD}& \textbf{MSR}\\
    \hline\hline
    Activities& 14 &  10 & 9& 18&16\\
    Subjects& 4 &  10 & 10&10 &10\\
    Videos& 60 &  200 & 215& 2160&320\\
    \hline
    \end{tabular}
    
    \label{tabel1}
\end{table}

\textbf{CAD-60}: This dataset includes 14 activities: \textit{rinsing mouth}, \textit{brushing teeth}, \textit{wearing contact lens}, \textit{talking on the phone}, \textit{drinking water}, \textit{opening pill container}, \textit{cooking (chopping)}, \textit{cooking (stirring)}, \textit{talking on couch}, \textit{relaxing on couch}, \textit{writing on whiteboard}, \textit{still (standing)}, \textit{working on computer} and \textit{random}. Each activity was performed by 4 subjects including one left-handed person. They were performed in 5 different environments: bathroom, bedroom, kitchen, living room, and office. It contains activities of cyclic nature such as \textit{brushing teeth} and activities with similar postures such as \textit{drinking water} and \textit{talking on the phone}.

\textbf{UTK}: There are 10 activities in this dataset: \textit{walk}, \textit{sit down}, \textit{stand up}, \textit{pick up}, \textit{carry}, \textit{throw}, \textit{push}, \textit{pull}, \textit{wave hands}, and \textit{clap hands}. These activities were performed by 10 subjects and repeated twice by each subject. The significant intra-class and viewpoint variations are the main challenges of this dataset.

\textbf{F3D}: This dataset includes 9 activities: \textit{wave}, \textit{drink from a bottle}, \textit{answer phone}, \textit{clap}, \textit{tight lace}, \textit{sit down}, \textit{stand up}, \textit{read watch}, and \textit{bow}. These activities were repeated twice or thrice by 10 subjects. The main challenge with this dataset is that the activities were performed at high speed. This provides a small number of frames for the algorithm to sample and learn from.

\textbf{KARD}: This dataset contains 2160 videos and consists of 18 activities. These activities were performed by 10 different people. The 18 activities are \textit{horizontal arm wave}, \textit{high arm wave}, \textit{two hand wave}, \textit{catch cap}, \textit{high throw}, \textit{draw X}, \textit{draw tick}, \textit{toss paper}, \textit{forward kick}, \textit{side kick}, \textit{take umbrella}, \textit{bend}, \textit{hand clap}, \textit{walk}, \textit{phone call}, \textit{drink}, \textit{sit down}, and \textit{stand up}. This dataset is challenging due to the large number of activities with high intra-class variation such as different individuals performing the same activity such as catch cap but in different ways, this makes learning of the same activity with large differences in movements difficult.

\textbf{MSR}: In this dataset there are 16 activities: \textit{drink}, \textit{eat}, \textit{read book}, \textit{call cellphone}, \textit{write on a paper}, \textit{use laptop}, \textit{use vacuum cleaner}, \textit{cheer up}, \textit{sit still}, \textit{toss paper}, \textit{play game}, \textit{lie down on sofa}, \textit{walk}, \textit{play guitar}, \textit{stand up}, and \textit{sit down}. There are 10 subjects, and each subject performed all activities in both standing and sitting positions. This makes the dataset challenging because the extracted features for both sitting and standing positions in each activity are different. Another challenge is data corruption. In some frames, the skeletal gesture structure suddenly collapses completely and lose their coherence and become meaningless.

\subsection{Method}
Through preliminary experiments, the best values for swarm size and the number of iterations were 20 and 50, respectively. The experiment was repeated 30 times and the average was obtained. The parameter settings of the HPGMK is summarized in Table~\ref{parameters}. The performance of our proposed method (HPGMK) was with three state-of-the-art (SOTA) methods for HAD incuding ENSC (elastic net subspace clustering), SSC (Sparse Subspace Clustering) and SC (Subspace Clustering)~\cite{paoletti2021subspace} and three recent deep clustering methods Deep Clustering Network (DCN)~\cite{yang2017towards}, Structural Deep Clustering Network (SDCN) ~\cite{van2020scan} and incomplete multi-view clustering via contrastive prediction (Completer)~\cite{lin2021completer}. In addition, we compared our method with conventional and well-known clustering methods including K-means Clustering (KM) and PSO. All parameters of each compared method, such as dimensions and numbers of layers, have been adjusted as described in their papers. K-means Clustering (KM) and PSO have been chosen for comparison as our proposed HPGMK is based on them. ENSC was found to be most similar to our work as an unsupervised algorithm requiring known cluster number while SSC and SC were the original algorithms that ENSC was based on. The three deep clustering methods were chosen to compare our method with the latest methods that use deep learning tools for clustering. To compare the performance of the methods, the accuracy metric (calculated based on~\cite{peng2019recursive}) was used. Moreover, F-score was used to show the performance of each method in categorizing each activity and the confusion between them was shown in the confusion matrix. The convergence test and clustering time of HPGMK were measured to evaluate the benefits of each component used in the HPGMK on its performance.
\begin{table}[h!]
\caption{Parameters setting used in the experiment.}
    \centering

    \begin{tabular}{c|c|c}
             \hline
    \textbf{Parameter}&\textbf{Description}&\textbf{Value}\\
    \hline\hline
     $c_{1_{max}},c_{2_{max}}$ &acceleration coefficients &2.5\\
     $c_{1_{min}}, c_{2_{min}}$&acceleration coefficients&0\\
     $w_{max}$&inertial weight& 0.9\\$t$&maximum number of iteration&50\\$T$& number of iteration
for mutation&10\\$np$& swarm size&20\\
    \hline
    \end{tabular}
    \label{parameters}
\end{table}

\subsection{Computation complexity analysis}
The Computation complexity of HPGMK in the initial stage (step 1 in Algorithm~\ref{HPGMK-algo}) is equal to $O(D.k.dim.np)$, where $D$ is the number of data points, $k$ is the number of clusters, $dim$ is the dimension of data points and $np$ is the population size of particles. The time complexity when particles are updated (step 2 to 9) is equal to $O(t.dim.T.np)$, where $t$ is maximum number of iteration and $T$ is the number of iteration for mutating~\textit{gbest}(t).
In Refining obtained centroids (step 11 to 15), the time complexity is equal to $O(log(D.k.dim))$. Therefore, the overall computation complexity of proposed algorithm is equal to $O(t.T.np.D.k.dim+log(D.k.dim))$. Table~\ref{computation} shows the comparison of the computation complexity of the well-known clustering algorithms including KM and BIRCH along with the state-of-the-art algorithms developed for clustering with the proposed algorithm. Based on the results, PSO-based clustering algorithms take substantially longer to execute than non-PSO-based clustering methods. 
But despite the low time complexity of KM and BIRCH clustering, their accuracy is very low and they need to be executed many times to reach the desired solution if they do not get stuck in the local optimum. HPSOK-means, MinMaxK-means, PSC-RCE, PSOLFK and PSOSCALFK  have around the same time complexity. However, we developed an algorithm, which has powerful global search capabilities due to increasing the variety of solutions during the execution of the algorithm by applying Guassina mutation and modifying the cluster using KM to increase the ability of PSO in local search. 

Distribution Method and Lévy Flight
\begin{table}[h!]
\caption{The computation complexity of HPGMK and eight related
algorithms.}
    \centering

    \begin{tabular}{cc}
             \hline
    \textbf{Algorithm}&\textbf{Computation Complexity}\\
    \hline\hline
    KM& $O(N.K.D.t)$ \\
    BIRCH& $O(N)$ \\
    HPSOK-means& $O(N.K.D.t)$ \\
    MinMaxK-means& $O(N.K.D.t)$  \\
    PSC-RCE& $O(N.K.D.n_{m}.t)$  \\
    PSOLFK& $O(N.K.D.t)$  \\
    PSOSCALFK& $O(N.K.D.t)$  \\
    KMM& $O((N.dim+m^{2}+m.k).t+m.dim)$  \\
    GLPSOK&$O(K.dim.(K.dim+N).t)$  \\
    HPGMK&$O(t.T.np.D.k.dim+log(D.k.dim))$  \\
    \hline
    \end{tabular}
    \label{computation}
\end{table}

\subsection{Results and discussion}
\begin{figure}
  \begin{center}
  \includegraphics[height=1.75 in,width=\textwidth]{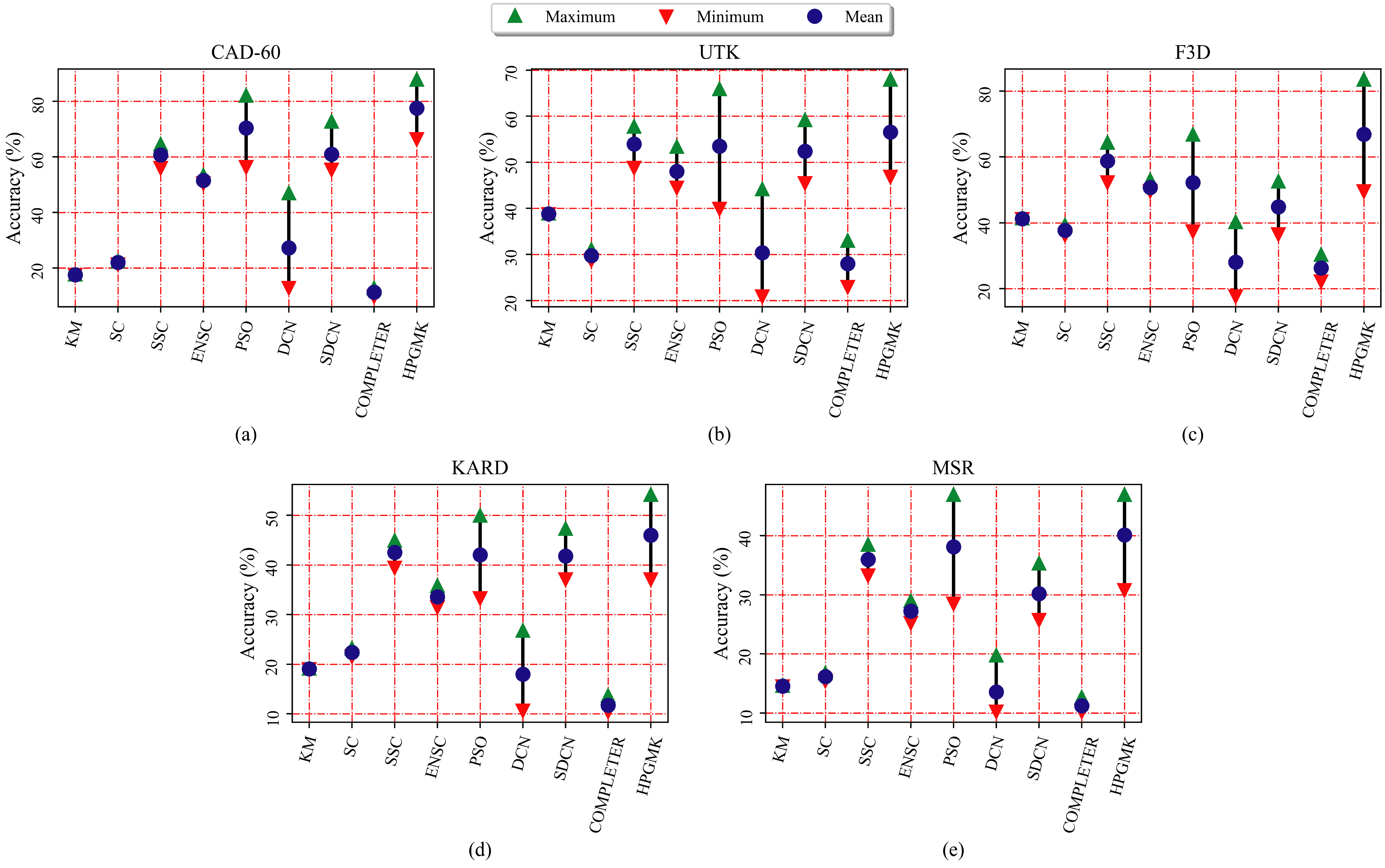}\\
  \caption{The average accuracy for all subjects in (a) CAD-60, (b) UTK, (c) F3D, (d) KARD and (e) MSR}\label{acc}
  \end{center}
\end{figure}

\begin{figure}
  \begin{center}
  \includegraphics[height=1.5 in,width=\textwidth]{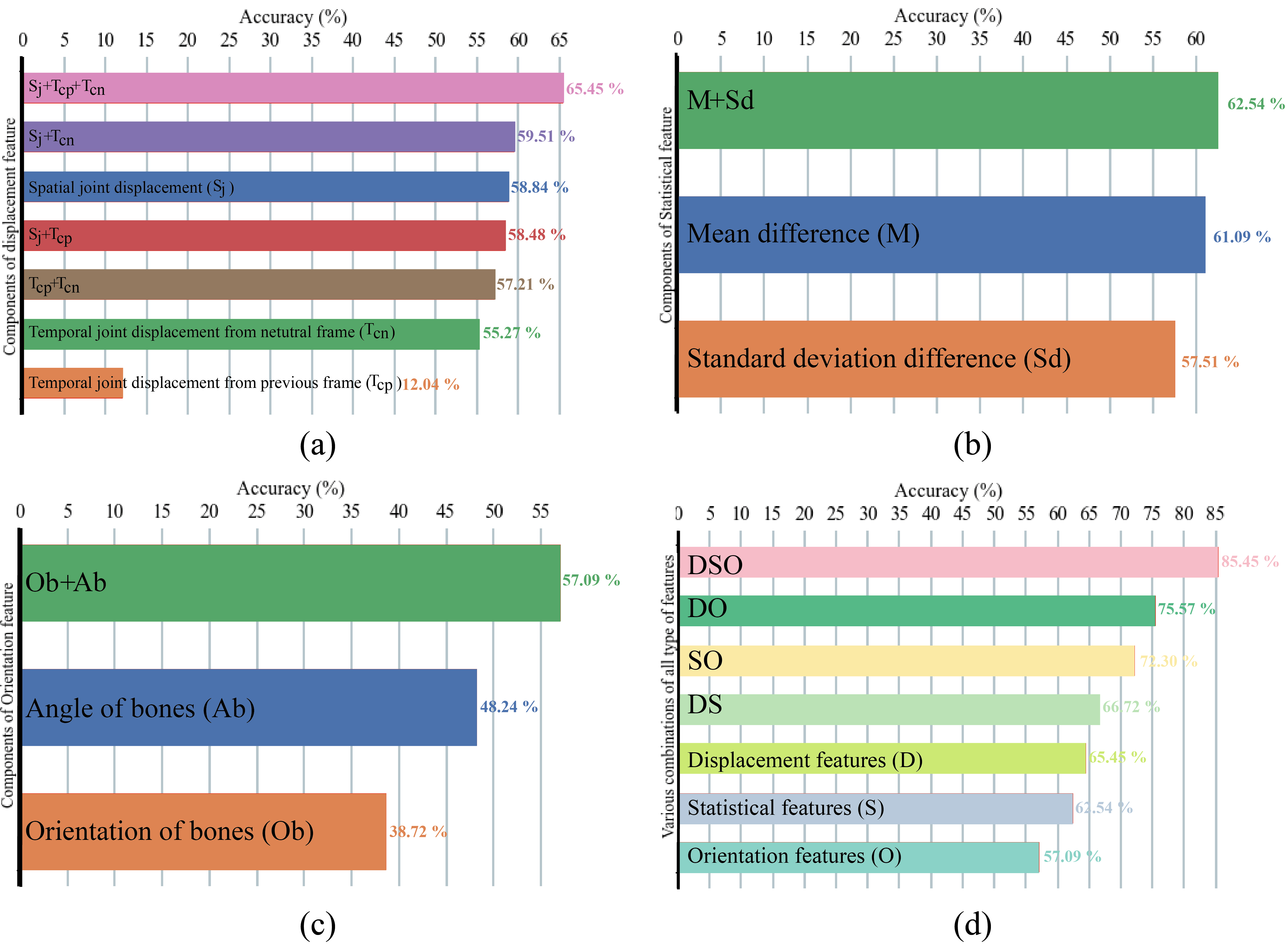} \\
  \caption{The effect of the each components of the each type of feature approach including (a) Displacement (D), (b) Statistical (S), (c) Orientation (O) features and (d) their various combinations together. Capital letters stand for different methods and putting these letters together means combining relevant methods.}\label{effect_of_features}
  \end{center}
\end{figure}

\begin{figure}
  \begin{center}
  \includegraphics[width=3 in]{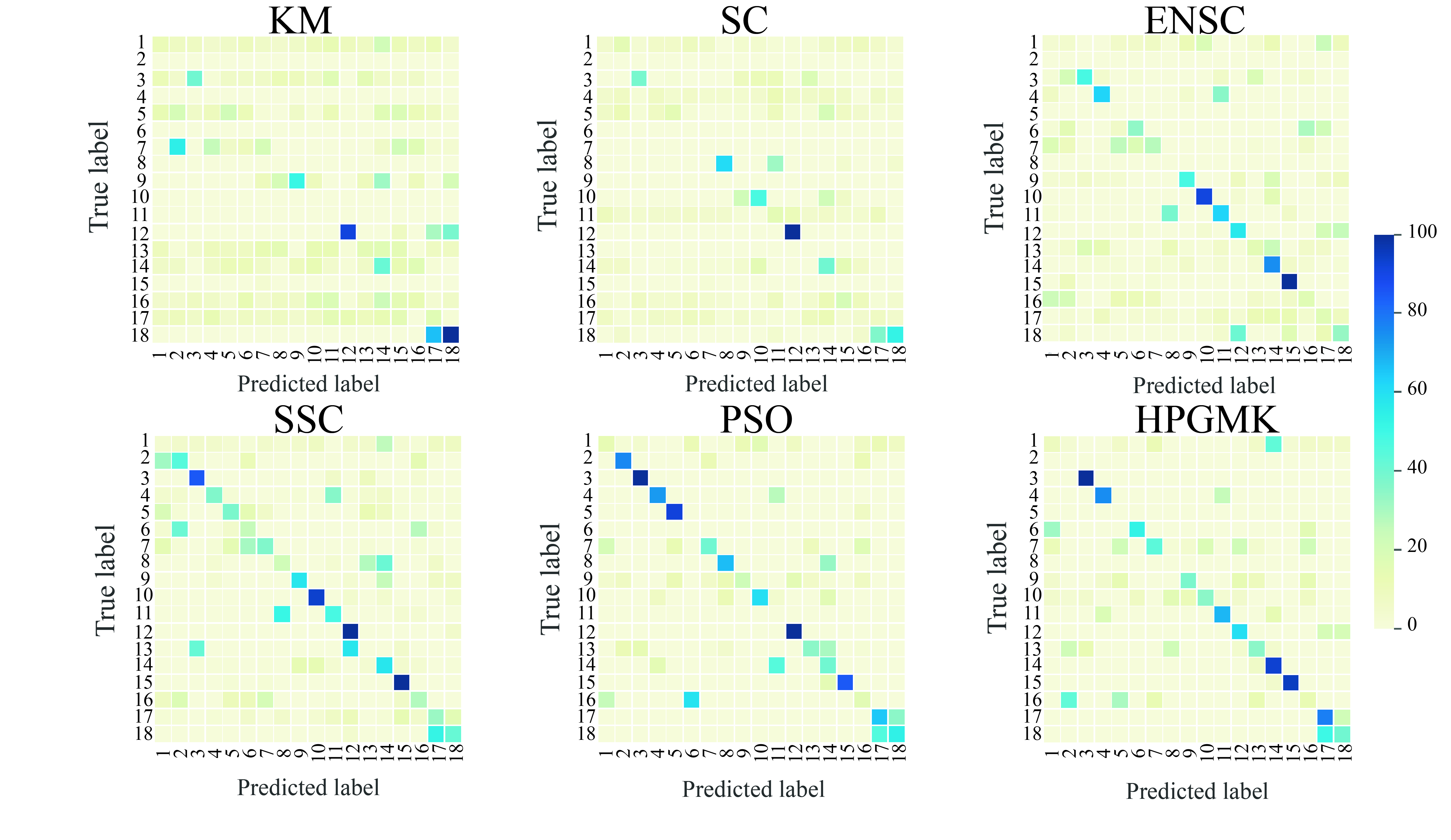}\\
  \caption{Comparison of confusion matrix of the result on subject 10 in KARD. Activity list: (1) Horizontal arm wave; (2) High arm wave; (3) Two hand wave; (4) Catch cap; (5) High throw; (6) Draw X; (7) Draw tick; (8) Toss paper; (9) Forward kick; (10) side kick; (11) Take umbrella; (12) Bend; (13) Hand clap; (14) Walk; (15) Phone call; (16) Drink; (17) Sit down; and (18) Stand up.}\label{kardcon}
  \end{center}
\end{figure}

Fig.~\ref{acc} shows the accuracy of HPGMK with the SOTA techniques for all subjects of each dataset based on the maximum, minimum and average accuracies. The average overall accuracy of the HPGMK was 77.53\,\% for CAD-60, 56.54\,\% for UTK, 66.84\,\% for F3D 46.02 \,\% for KARD and, 40.12 \,\% for MSR. As seen in Fig.~\ref{acc}, HPGMK has the best performance in terms of maximum and average accuracy in all datasets. This shows the effectiveness of the HPGMK for human activity discovery. By utilizing the Gaussian mutation and KM along with PSO, our approach brings performance improvement compared to the other methods. ENSC and SSC, which are subspace clustering algorithms, do not use an efficient search strategy~\cite{lu2011particle}. In these methods, there is no strategy for maintaining the balance between exploitation and exploration in their search. Moreover, Parameters are required to be set and finding the right values for them is tricky and complex such as size of subspace~\cite{agarwal2021meta}. In contrast, HPGMK are not dependent to parameters like SSC and ENSC and has several strategies for searching. First, it used the PSO to search in large space area by using several particles as potential solutions. To promote diversity, Gaussian mutation is used. KM is also used to search in a small area of the  global best solution to refine the obtained centroids from PSO. These search strategies, enable HPGMK has a relatively good performance compared to the SSC and ENSC. Moreover, HPGMK has performed better than the methods that use deep learning. In HPGMK, both spatial features from each 3D skeleton frame and temporal features from sequences along with Orientation and statistical information are extracted. However, in the deep clustering this information is ignored. On the other hand, unlike deep clustering methods that have used shallow clustering, HPGMK has different search strategies for exploration and exploitation to determine better clusters.
Fig.~\ref{effect_of_features}(a) to (c) show the effect of each component of each type of the proposed hybrid feature extraction method based on the discovery accuracy of the activities performed by subject one in the CAD-60 dataset. Percentages represent the discovery accuracy using the different combination of features and each piece of graphs shows the ratio of the impact of one component in discovery improvement to the rest of the other components in each type of feature. This ratio is obtained based on dividing \textit{discovery accuracy obtained by one of the components from a feature type} by \textit{summation of discovery accuracy obtained by all components of that feature type}. Overall, in three feature extraction methods comprising displacement, statistical, and orientation features, when all their components are combined, the discovery accuracy significantly increased and obtained 65.45\,\%, 62.54\,\%, and 57.09\,\% respectively. By contrast, if each component of the feature extraction methods is used alone without considering other components in the feature extraction, the accuracy of discovery decreases. Fig.~\ref{effect_of_features}(d) shows the effect of different combinations of each of the feature extraction methods. The size of each circle indicates the effectiveness of the features. Based on the obtained results, it is shown that the highest detection accuracy of 85.45\,\% is obtained by combining all the three methods. It indicates that in order to better differentiate between activities, it is necessary to extract features from different aspects of activities.
\begin{figure}
  \begin{center}
  \includegraphics[height=2 in,width=\textwidth]{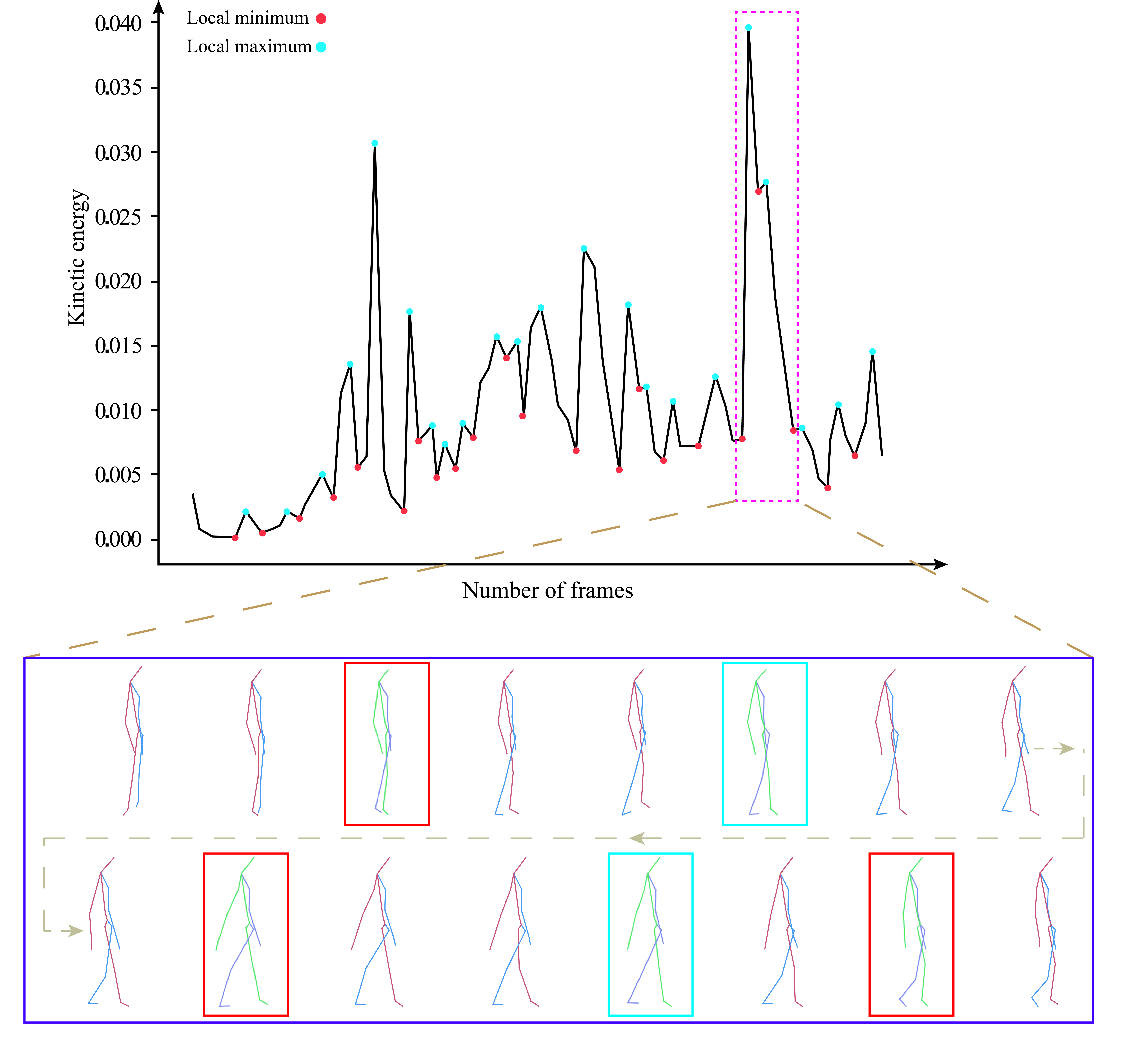}\\
  \caption{Illustration of the effect of selecting the keyframe based on kinetic on "walking" in MSR where distinct frames compared to adjacent frames are selected from similar frames. A window of frames is shown with a few selected frames based on the local maximum (blue) and local minimum (red)  energy.}\label{Keyframes_results}
  \end{center}
\end{figure}
Fig.~\ref{Keyframes_results} shows the selected keyframes from walking activity in MSR. As shown, there are a lot of frames with high similarity that by extracting their features, not only do not help to improve discovery but also increase the computational complexity and increase the overlap between other activities because these gestures occur in other activities. However, using local maximum and minimum kinetic energy can find representative frames and reduce complexity.
Looking at a window of frames, we can see the selected frames based on the maximum and minimum local energy value. The selected frames show the most differentiation to display the activity sequence. It is worth mentioning that selecting keyframes maintains the order of the activity.\\
To show activity discovery performance, confusion matrix in on
subject 10 in KARD was shown in Fig.~\ref{kardcon} for different methods. This figure demonstrates that HPGMK produced distinctively meaningful clusters. In this figure, cluster overlapping appears relatively high in the other methods especially KM and SC in all datasets. There were a lot of overlapping due to the large number of activities that are very similar including similar body gestures and similar hand movement. For instance, activity overlapping were seen between \textit{High arm wave} and \textit{High throw} with \textit{Drink}.

\begin{figure}
  \begin{center}
  \includegraphics[width=3 in]{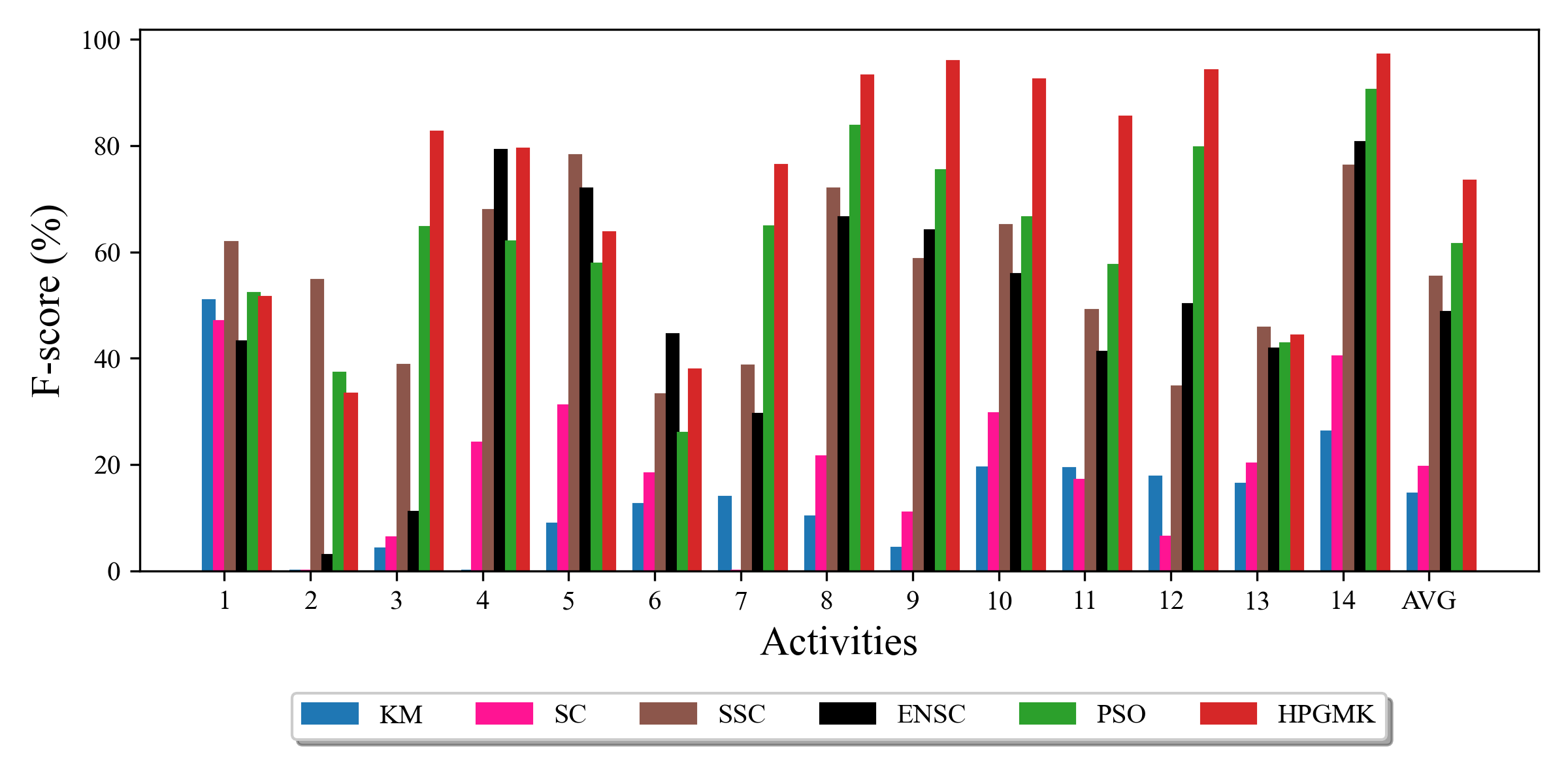}\\
  \caption{The average F-score for all subjects in CAD-60. Activity list: (1) Brushing teeth; (2) Cooking (chopping); (3) Rinsing mouth with water; (4) Still(standing); (5) Taking on the couch; (6) Talking on the phone; (7) Wearing contact lenses; (8) Working on computer; (9) Writing on whiteboard; (10) Drinking water; (11) Cooking (stirring); (12) Opening pill container; (13) Random; and (14) Relaxing on couch. AVG is the average F-score for all activities.}\label{cad_fscore}
  \end{center}
\end{figure}
\begin{figure}
  \begin{center}
  \includegraphics[width=\textwidth]{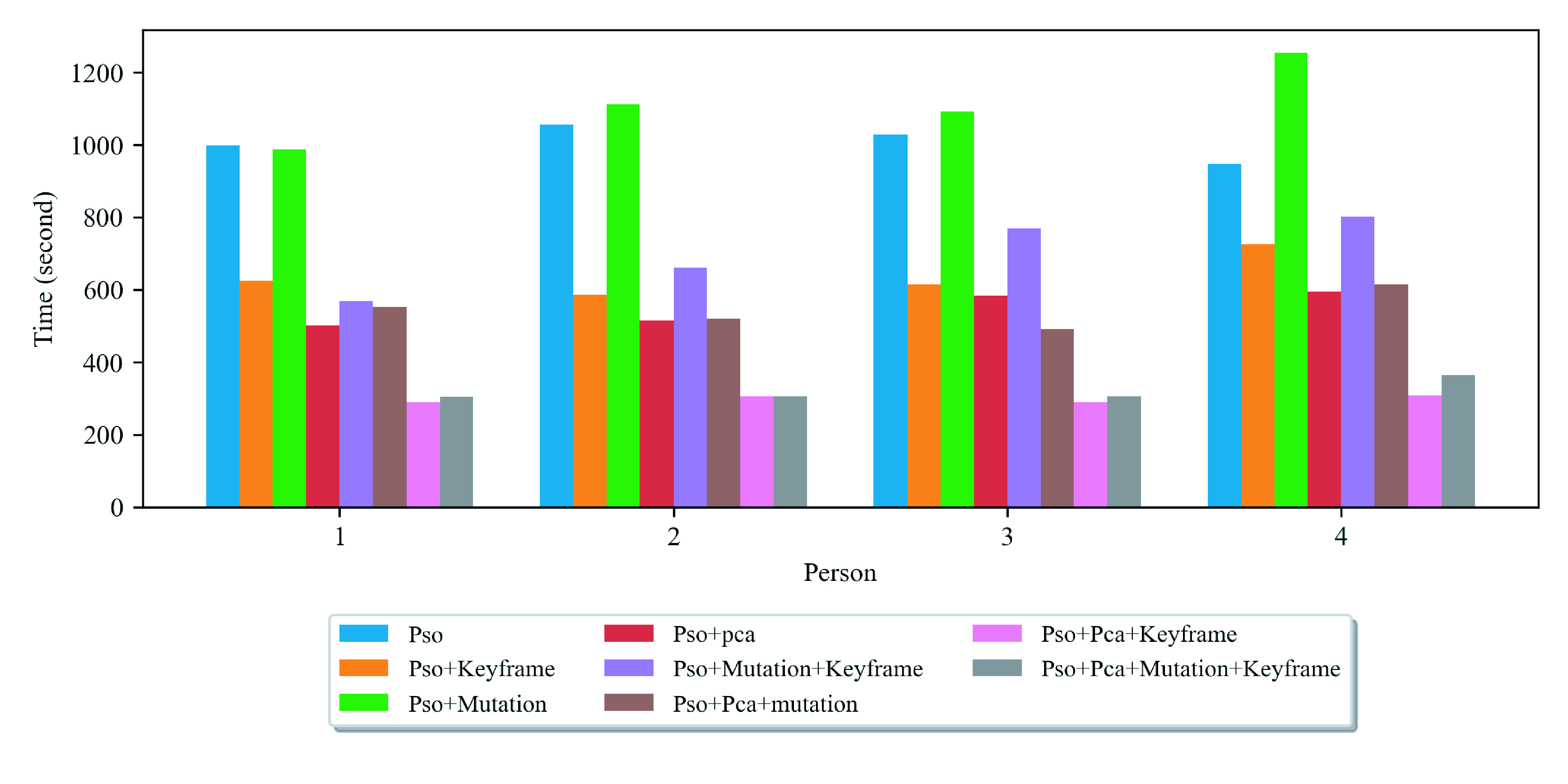}\\
  \caption{The comparison of clustering time of different components of proposed algorithm on all subjects in CAD-60.}\label{time}
  \end{center}
\end{figure}

\begin{figure}
  \begin{center}
  \includegraphics[width=\textwidth]{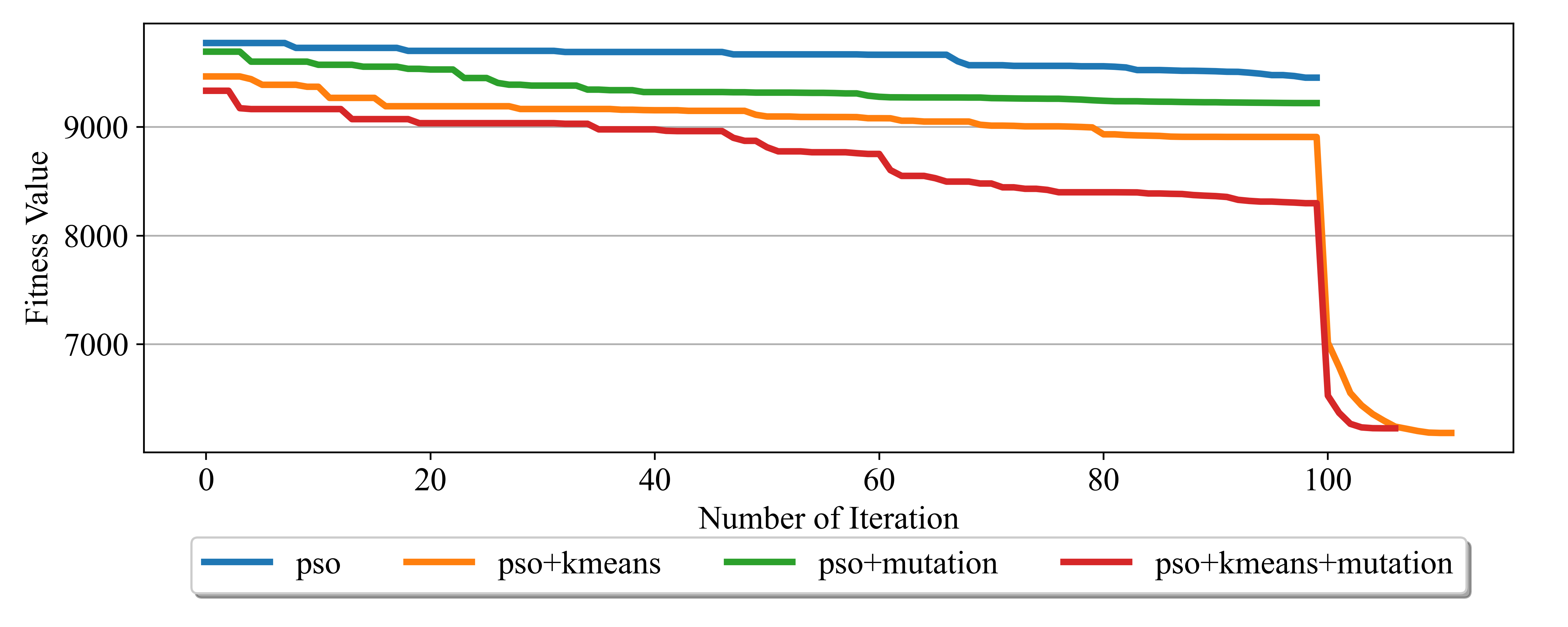}\\
  \caption{The comparison of convergence in subject 1 of CAD-60.}\label{convergence}
  \end{center}
\end{figure}

Fig.~\ref{cad_fscore} shows the average F-score for each activity of all subjects in CAD-60. By examining the average F-scores in most activities, it shows HPGMK outperforming other methods and achieved slightly under 75\,\% on CAD-60 in average for all activities. Although F-score was high in most of the activities for HPGMK compared to other methods, \textit{cooking (chopping)} and \textit{talking on the phone} were discovered with low F-scores. This was due to the high similarity between \textit{cooking (chopping)} with \textit{cooking (stirring)} and \textit{talking on the phone} with \textit{wearing contact lenses}. 

Fig.~\ref{time} shows the average clustering time of the different combinations of components used in the proposed algorithm in milliseconds on subject 1 in CAD-60. This experiment evaluates the impact of the different components in the proposed HPGMK algorithm. As can be seen, the clustering time of the proposed algorithm (red line) was relatively low. The reason is that our approach has benefited from dimension reduction methods, including PCA and keyframe selector. 
Fig.~\ref{convergence} indicates the effect of each component of HPGMK on convergence rate. As it is indicated, each combination has a different convergence rate. However, using all components enables the proposed method to achieve the best convergence compared to the other combinations. 
It is also confirmed from Fig.~\ref{convergence} that by combining both KM and PSO algorithms, the convergence speed has increased.
Table~\ref{kurooss} indicates a significant difference between the HPGMK and KM, HPGMK and PSO, HPGMK and SSC, HPGMK and ENSC and HPGMK and SDCN through 30 independent runs using the Kruskal–Wallis test (p-value). Since the p-value of almost all of the datasets is less than 0.05 (significance level) with the 95\,\% confidence intervals for each median, we reject the null hypothesis, and conclude there is a significant difference between the proposed method with KM, PSO, SSC, ENSC, and SDCN. In cases where the p-value is higher than the significant level (these cases are specified with${}^\star$ in Table~\ref{kurooss}), there is not enough evidence to reject the null hypothesis.
\begin{table}[h!]
\caption{Comparison of Kruskal-Wallis test (p-value) between HPGMK and
KM, HPGMK and PSO, HPGMK and SSC, HPGMK and ENSC and HPGMK and SDCN through 30 independent runs}
\begin{threeparttable}
    \centering

\scalebox{0.53}{
    \begin{tabular}{c|c|c|c|c|c}
             \hline
             Datasets & HPGMK vs KM &HPGMK vs PSO  & HPGMK vs SDCN & HPGMK vs ENSC & HPGMK vs SSC\\
    \hline\hline
  CAD-60 &$\approx$0 &0.00048& $\approx$0 &$\approx$0\\
  UTK & $\approx$0 &0.00014& 0.02369${}^\star$&$\approx$0&0.03325${}^\star$\\
  F3D &$\approx$0&$\approx$0&$\approx$0 &$\approx$0&$\approx$0 \\
  KARD &$\approx$0&0.00544&0.023696${}^\star$&$\approx$0&0.39117${}^\star$\\
  MSR &$\approx$0 &0.88246${}^\star$& $\approx$0&$\approx$0&0.018736${}^\star$\\
    \hline
    \end{tabular}}
    
    \begin{tablenotes}
        \small
      \item ${}^\star$ p-value $>$ 0.05: The differences between the medians are not \\statistically significant.
        \end{tablenotes}
     \end{threeparttable}
    \label{kurooss}
\end{table}

\section{Conclusion}\label{Conclusion}
Most of the proposed HAR frameworks are supervised or semi-supervised, making them unusable in real-world situations due to a lack of access to the ground truth. In this paper, a Hybrid Particle Swarm Optimization with Gaussian Mutation and k-means (HPGMK) approach was proposed to solve human activity discovery on skeleton-based data with no prior knowledge of the label of the activities in the data. Five different datasets were used to assess the performance of the method. The results obtained have shown that HPGMK achieved an average overall accuracy of 77.53\,\%, 56.54\,\%, 66.84\,\%, 46.02\,\%, and 40.12\,\% in datasets CAD-60, UTK, F3D, KARD, and MSR, respectively and validate the superiority of HPGMK over other methods compared. In activities with high intra-class variation, corrupted data and the same activity performed in sitting and standing positions, HPGMK has performed better in activity discovery than other SOTA methods. \\
We have examined the impact of each feature used. It was found that the simultaneous combination of features together further improves the results. The impact of the different components in the proposed algorithm has shown that Gaussian mutation has evolved particles to improve search algorithm and k-means has increased discovery efficiency and improved the convergence rate.  \\  
This work paves the way toward implementing fully unsupervised human activity discovery in practical applications using skeleton-based data. There are various factors in the proposed methods that need to be addressed to develop an effective activity discovery algorithm. One factor is the number of clusters that were pre-configured in the proposed algorithm. The HPGMK needs to be further extended to automatically address human activity discovery by estimating the number of activities by itself. Another factor is detecting outlier or noisy data. Outliers shift the cluster centers towards themselves, thus affecting optimal cluster formation. Using outlier detection methods in HPGMK to reject outliers will be beneficial. The manual procedure used to build the PSO structure in the suggested technique was based on the information and expertise that was gained. It takes a long time to manually introduce changes because of trial-and-error, which makes it challenging to thoroughly explore all potential algorithm setups. A potential future study to address these concerns is automating the suggested method's setup to make it more effective in handling various circumstances and datasets.

\section*{Acknowledgment}

This work was supported by Grant UBD/RSCH/1.11/ FICBF(b)/2019/001 from Universiti Brunei Darussalam.


%





\ifCLASSOPTIONcaptionsoff
  \newpage
\fi





\bibliographystyle{IEEEtran}
\bibliography{IEEEabrv,Bibliography}

\vfill

\end{document}